\documentclass{article}

\pdfoutput=1
\PassOptionsToPackage{numbers, compress}{natbib}

\usepackage{multirow}
\usepackage{adjustbox}
\usepackage{subcaption}
\usepackage{enumitem}

\usepackage[preprint]{style}

\usepackage[utf8]{inputenc} 
\usepackage[T1]{fontenc}    
\usepackage[colorlinks,linkcolor=red,citecolor=blue,urlcolor=magenta,bookmarks=true]{hyperref}       
\usepackage{url}            
\usepackage{booktabs}       
\usepackage{amsfonts}       
\usepackage{nicefrac}       
\usepackage{microtype}      
\usepackage{xcolor}         
\usepackage{subcaption}
\usepackage{comment}
\usepackage{floatrow}
\newfloatcommand{capbtabbox}{table}[][\FBwidth]
\usepackage{array}
\usepackage{multirow}
\usepackage{wrapfig}
\usepackage{bm}
\usepackage{amsmath}
\usepackage{color}
\usepackage{diagbox}
\usepackage{amssymb}
\usepackage{setspace}
\usepackage{pifont}
\usepackage{adjustbox}

\vspace{-5mm}
\title{\vspace{-3mm}Playground v3: Improving Text-to-Image Alignment with Deep-Fusion Large Language Models\vspace{-3mm}}

\author{
  Bingchen Liu
  \quad  Ehsan Akhgari
  \quad Alexander Visheratin
  \quad Aleks Kamko 
  \quad \textbf{Linmiao Xu} \\
  \quad \textbf{Shivam Shrirao} 
  \quad \textbf{Chase Lambert} 
  \quad \textbf{Joao Souza} 
  \quad \textbf{Suhail Doshi}
  \quad \textbf{Daiqing Li}
  \vspace{3mm}\\
  Playground Research
  \vspace{2mm}
}

\begin{document}

\maketitle

\maketitle
\begin{center}
\vspace{-9mm}
    \centering
  	\captionsetup{type=figure}
	\includegraphics[width=0.9\textwidth]{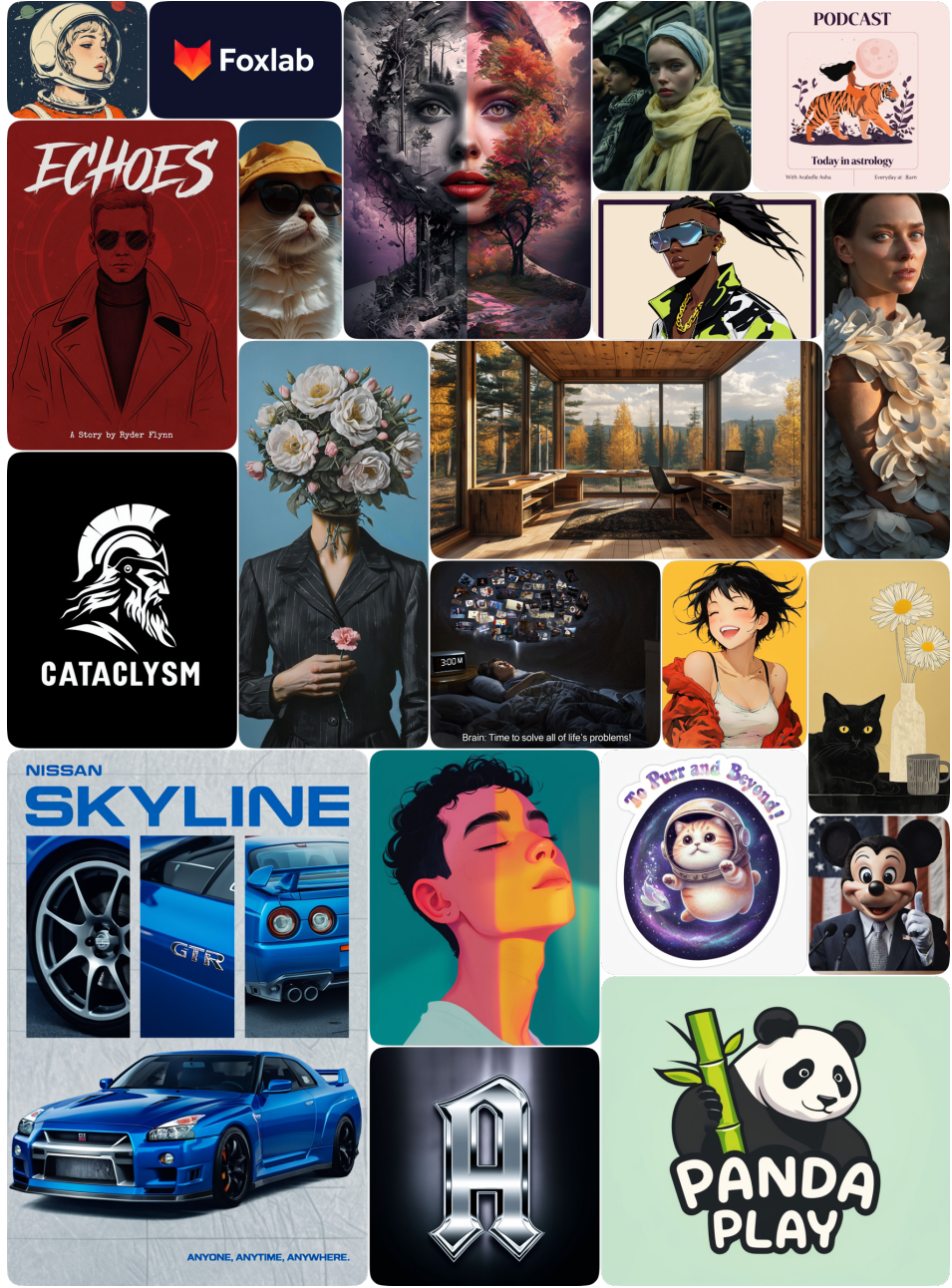}
 
\captionof{figure}{\small {Photo-realistic people, diverse scenes, accurate text, and artistic designs samples from Playground v3.}}
\label{fig:teaser}
\end{center}

\newpage

\begin{abstract}
We introduce Playground v3 (PGv3), our latest text-to-image model that achieves state-of-the-art (SoTA) performance across multiple testing benchmarks, excels in graphic design abilities and introduces new capabilities. Unlike traditional text-to-image generative models that rely on pre-trained language models like T5 or CLIP text encoders, our approach fully integrates Large Language Models (LLMs) with a novel structure that leverages text conditions exclusively from a decoder-only LLM. Additionally, to enhance image captioning quality—we developed an in-house captioner, capable of generating captions with varying levels of detail, enriching the diversity of text structures. We also introduce a new benchmark \textit{CapsBench} to evaluate detailed image captioning performance. Experimental results demonstrate that PGv3 excels in text prompt adherence, complex reasoning, and accurate text rendering. User preference studies indicate the super-human graphic design ability of our model for common design applications, such as \textit{stickers}, \textit{posters}, and \textit{logo} designs. Furthermore, PGv3 introduces new capabilities, including precise RGB color control and robust multilingual understanding.

\end{abstract}
\section{Introduction}

Great progress has been made in the text-to-image generative models since last year \cite{nichol2021glide,ramesh2022hierarchical,rombach2022highresolution,esser2024scalingrectifiedflowtransformers,saharia2022photorealistictexttoimagediffusionmodels}. The model architecture for large-scale text-to-image model is shifting from a traditional UNet-based\cite{nichol2021glide} model to a transformer-based model\cite{peebles2023scalable} due to its scalability and simplicity. In this work, we built on this progress to develop a new DiT-based diffusion model and scale it to 24B parameters. We further propose a Deep-Fusion architecture to leverage the knowledge of the modern decoder-only large-language models \cite{dubey2024llama} for the task of text-to-image generation.

In section~\ref{sec:method} we describe the simplified model architecture, noise scheduling, and the new Variational Autoencoder (VAE). Our novel text-to-image model structure features a deep integration of an LLM, fully leveraging the LLM’s internal prompt understanding to achieve state-of-the-art prompt-following performance. Section~\ref{sec:training} discusses training details, including the use of multi-level captions and model merging during the post-training phase. Section~\ref{sec:captioner} introduces our in-house captioner and a new captioning benchmark \textit{CapsBench}. 

In section~\ref{sec:qualitative}, we present qualitative examples of our model’s capabilities across five key aspects: photo-realism, prompt-following, text-rendering, RGB color control, and multilingual understanding. Most of the images displayed are generated from our evaluation testing prompt set, which was curated through our human-in-the-loop pipeline and excluded from our training set.

In section~\ref{sec: image-eval}, we quantitatively evaluate our image models across four categories. Notably, we demonstrate that PGv3 exhibits superior graphic design capabilities, often surpassing even real human designers. We address image-text alignment and reasoning using the LLM-aided DPG-benchmark \cite{hu2024ella}. To further validate this, we developed an in-house version of the DPG-benchmark with more complex testing prompts and used GPT-4o as the VQA evaluator,; this new benchmark  demonstrates PGv3's SoTA prompt-following performance. Additionally, we briefly showcase the reconstruction quality of our new VAE. Finally, we assess image quality using standard benchmarks such as ImageNet and MSCOCO. For our in-house captioner model, details and results of the \textit{CapsBench} are discussed in section~\ref{sec:caption_eval}.
\section{Methods}
\label{sec:method}
\subsection{Model}
Playground-v3 (PGv3) is a Latent Diffusion Model (LDM) \cite{rombach2022highresolution} trained using the EDM \cite{karras2022elucidating} formulation. Like other models such as DALL-E 3 \cite{betker2023improving}, Imagen 2 \cite{saharia2022photorealistictexttoimagediffusionmodels}, and Stable Diffusion 3 \cite{esser2024scalingrectifiedflowtransformers}, PGv3 is designed for text-to-image (t2i) generation tasks. However, it distinguishes itself through a novel model structure, diverging from the commonly-used text encoders like T5 \cite{raffel2020exploring} or CLIP \cite{radford2021learning}, which we consider to be insufficient. Instead, PGv3 fully integrates a large language model (LLM), specifically Llama3-8B \cite{dubey2024llama}, to enhance its capabilities in prompt understanding and following.

There have been a lot of research efforts aimed at leveraging LLMs to improve t2i models' prompt-following abilities. These efforts typically involve either using LLMs as the text encoder to replace the commonly-used T5 and CLIP, or employing LLMs to adapt or revise prompts for a pretrained T5/CLIP-based t2i model \cite{he2024mars,song2024moma,qin2024diffusiongpt,hu2024ella,liu2024llm4gen,behnamghader2024llm2vec}. In PGv3, we adopt the former approach, i.e. we don't use any NLP encoders like T5 and CLIP, and solely rely on Llama3-8B to provide text conditioning to our diffusion model, from the initial stage of the model training.

\subsubsection{Text Encoder}
We take a notably different approach to using a pre-trained text model. The standard practice has been to use the output from the final layer of the T5 encoder or CLIP text encoder, with some research also finding value in incorporating the penultimate layer’s output. It is well-established that each layer within a transformer model captures different representations, containing varying levels of word-level and sentence-level information \cite{durrani2020analyzing,dar2022analyzing,Rombach_2022_CVPR}. However, we found that selecting the optimal layers for conditioning the t2i model can be cumbersome, particularly when using a decoder-style LLM, which arguably possesses superior generative power due to its more complex internal representations.

We believe that the continuity of information flow through every layer of the LLM is what enables its generative power and that the knowledge within the LLM spans across all its layers, rather than being encapsulated by the output of any single layer. Motivated by this, we designed our t2i model structure to replicate all the transformer blocks of an LLM. This design allows our model to take the hidden embedding outputs from each corresponding layer from LLM as conditioning at each respective layer of our t2i model. We argue that this approach fully leverages the complete "thinking process" of the LLM, guiding our t2i model to mirror the LLM’s reasoning and generative process. As a result, our model achieves unparalleled prompt-following and prompt-coherence abilities when generating images.

\subsubsection{Model Structure}
\begin{figure}
	\centering
	\includegraphics[width=13cm]{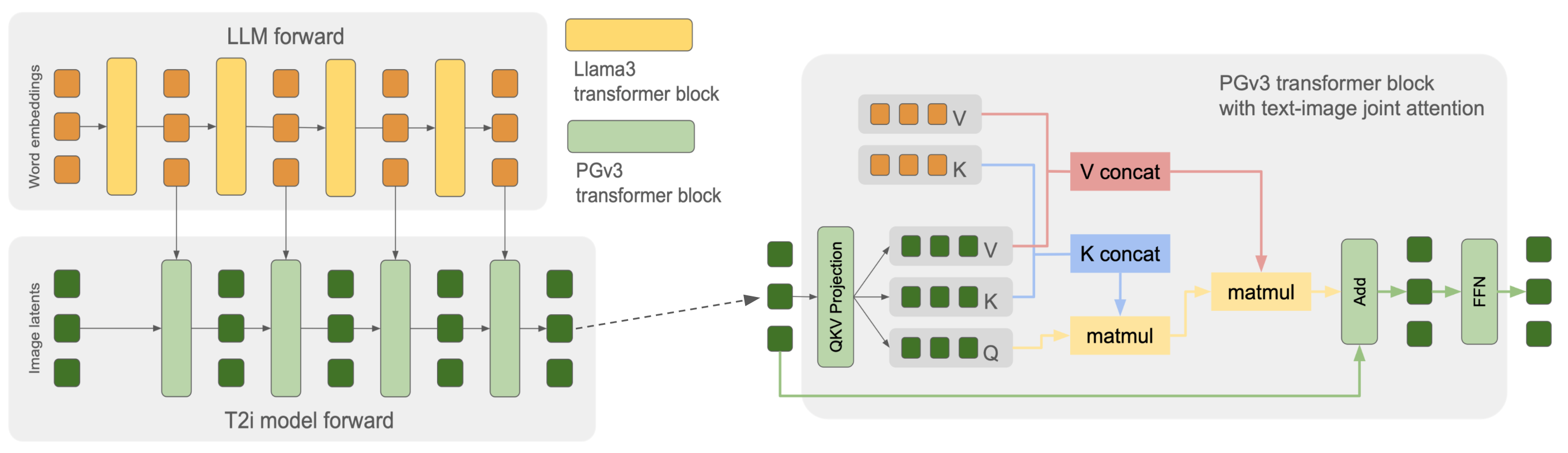}
	\caption{PGv3 model structure overview simplified}
	\label{fig:model_structure}
\end{figure}
PGv3 employs a DiT-style \cite{peebles2023scalable} model structure. Each transformer block in our image model is set up as identical to the corresponding block in the LLM we utilize, Llama3-8B in this case, including matching parameters such as hidden dimension size, number of attention heads, and attention heads dimensions. 
We trained only the image model component, leaving the LLM untouched. Note that during the diffusion sampling process, the LLM part of the network only needs to run once to generate all the intermediate hidden embeddings required.

The simplified model block is illustrated in Fig.\ref{fig:model_structure}. Each transformer block consists of only one attention layer and one feed-forward layer, mirroring the structure of Llama3. Unlike most traditional CNN-based diffusion models \cite{ramesh2022hierarchical,Rombach_2022_CVPR}, which typically separate self-attention on image features from cross-attention between image and text features, we performed a single joint attention operation. In this approach, the image query attends to a concatenation of image keys and text keys, allowing it to extract relevant features from a combined pool of image and text values. This joint attention design is primarily motivated by the desire to reduce computational costs and inference time.

Although we finalized our model design and began training before the publication of the SD3 paper \cite{esser2024scalingrectifiedflowtransformers}, we have observed that the PGv3 transformer block shares a similar spirit with the MMDiT proposed in SD3. When considering the corresponding LLM block and image block as a single layer in PGv3, our approach reflects a similar philosophy to MMDiT—enabling integrated and continuous interaction between image and text features across all model layers. However, our design is more minimalist and leverages the text encoder more thoroughly.

There have been many research efforts aimed at exploring better model structures, such as \cite{li2024hunyuan,Karras2024edm2,chen2023pixart}. We experimented with several structural tweaks for PGv3 but did not observe significant benefits from most of them. Here we report on some designs that we found useful and incorporated into our final model design.

\textbf{U-Net skip connections between transformer blocks.}
We applied U-Net \cite{ronneberger2015u} skip connection across all the image blocks, as previously tried in U-DiT\cite{tian2024u}.

\textbf{Token down-sampling at middle layers.}
Among the 32 layers, we reduced the sequence length of the image keys and values by four times in middle layers
making the whole network resembles a traditional convolution U-Net with only one level of down sampling. This adjustment slightly speeds up both training and inference time, and we did not observe any performance degradation as a result.

\textbf{Positional embedding.}
We used the traditional Rotary Position Embedding (RoPE) \cite{su2024roformer}, similar to what is used in Llama3. Since we work with 2-dimensional image features, we employed a 2D version of RoPE. Several prior works have explored variations of 2D-RoPE; in our study, we primarily experimented with two types, which we refer to as ``interpolating-PE" and ``expand-PE."

The ``interpolating-PE" approach keeps the start and end positional IDs fixed, regardless of sequence length, and interpolates the positional IDs in between—an idea employed by models like SD3 and others. The ``expand-PE" method, on the other hand, increases the positional ID proportionally with sequence length, without applying any tricks or normalization (i.e., vanilla positional embedding).

We found that the ``interpolating-PE" had a significant downside: it caused the model to overfit heavily on training resolutions and failed to generalize to unseen aspect ratios. In contrast, the ``expand-PE" method performed well, showing no signs of resolution overfitting. Therefore, we opted to use the traditional positional embedding in the end.

\subsubsection{New VAE}
The Variational Autoencoder (VAE) \cite{kingma2013auto} used in Latent Diffusion Models (LDM) plays a crucial role in determining the model's upper bound for fine-grained image quality. In the work of Emu \cite{dai2023emu}, they increased the VAE latent channels from 4 to 16, which resulted in an enhanced ability to synthesize finer details, especially for smaller faces and words. Following this approach, we trained our in-house VAE with 16 channels. Additionally, instead of training solely at a \(256*256\) resolution, we extended the training to a \(512*512\) resolution, which further enhanced reconstruction performance.

We attempted to incorporate the frequency method proposed in Emu's VAE training but did not observe any performance improvement. Additionally, we explored more aggressive VAE modifications, including using pretrained image encoders like the CLIP image encoder or the DINOv2 encoder \cite{oquab2023dinov2}. Unfortunately, although these approaches yielded VAEs with comparable reconstruction abilities, the latent spaces generated by these pretrained image encoders, whether frozen or fine-tuned, resulted in unstable diffusion training, therefore we did not pursue this exploration further.

\begin{figure}
	\centering
	\includegraphics[width=13cm]{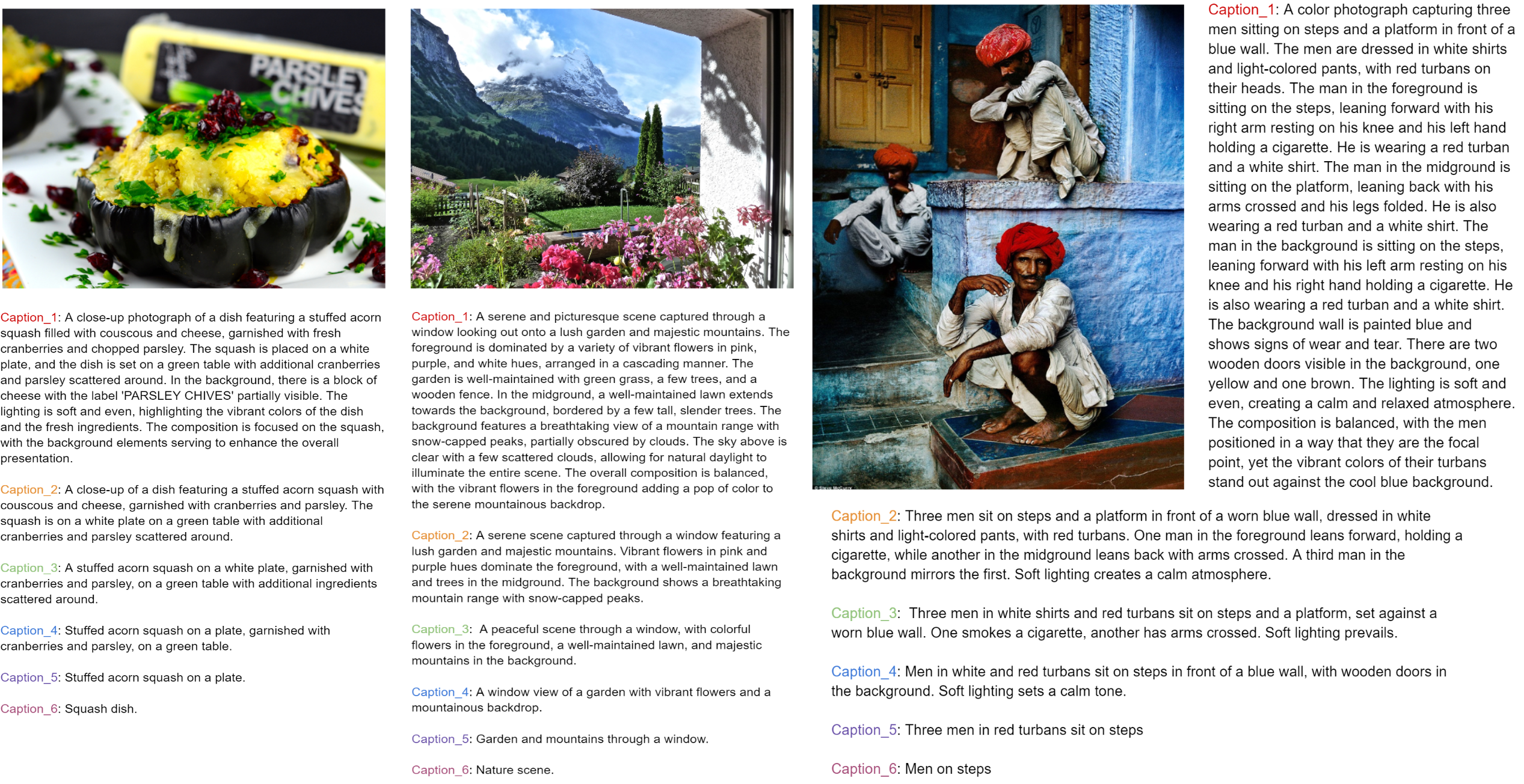}
	\caption{Multi-level caption example, randomly sampled from our training data set.}
	\label{fig:caption_example}
\end{figure}
\vspace{-3mm}
\section{Model Training}
\label{sec:training}
\textbf{Noise schedule}.
Despite the recent rise of flow-matching models \cite{lipman2022flow,liu2022flow,esser2024scalingrectifiedflowtransformers}, we continue to use the EDM schedule \cite{karras2022elucidating}, as we did in PGv2.5. There is no specific reason for not transitioning to flow-matching; we simply have not encountered any performance limitations with the EDM schedule in our new PGv3 model structure. In the mean time, we did experiment with the dynamic timestamp weighting technique proposed in EDM2 \cite{Karras2024edm2}, which is tailored to the EDM schedule. Unfortunately, we did not observe any significant performance improvements. We suspect this may be due to differences in model size and data, as our training scale is much larger and more diverse than the ImageNet training setup used in EDM2.

\textbf{Multiple Aspect-Ratio Support}.
Following our previous work in PGv2.5 \cite{li2024playground}, we adopted the multi-aspect ratio training strategy. We started with square image training at \(256*256\) low-resolution and used the online bucketing strategy at the higher resolution \(512*512\) and \(1024*1024\) pixel scales.

\textbf{Multi-Level Captions per Image}.
We developed an in-house Vision Large Language Model (VLM) captioning system capable of generating highly detailed descriptions of images, including small objects, lighting, style, and visual effects. We ran benchmarks on this in-house captioner (PG Captioner) against other state-of-the-art systems, such as GPT-4o. Our results show that our model outperforms others across multiple image-specific categories, which will be discussed in detail later.

Previous studies \cite{betker2023improving, esser2024scalingrectifiedflowtransformers} have proposed re-captioning datasets and using machine-generated captions as text conditions to enhance model performance. In our work, we further improved these captioning conditions by generating multi-level captions to reduce dataset bias and prevent model overfitting. As shown in Fig.\ref{fig:caption_example}, for each image, we synthesized captions of six different lengths, ranging from fine-grained details to coarse concepts. During training, we randomly sampled one of these six captions for each image at different iterations.

\begin{figure}
	\centering
	\includegraphics[width=13.5cm]{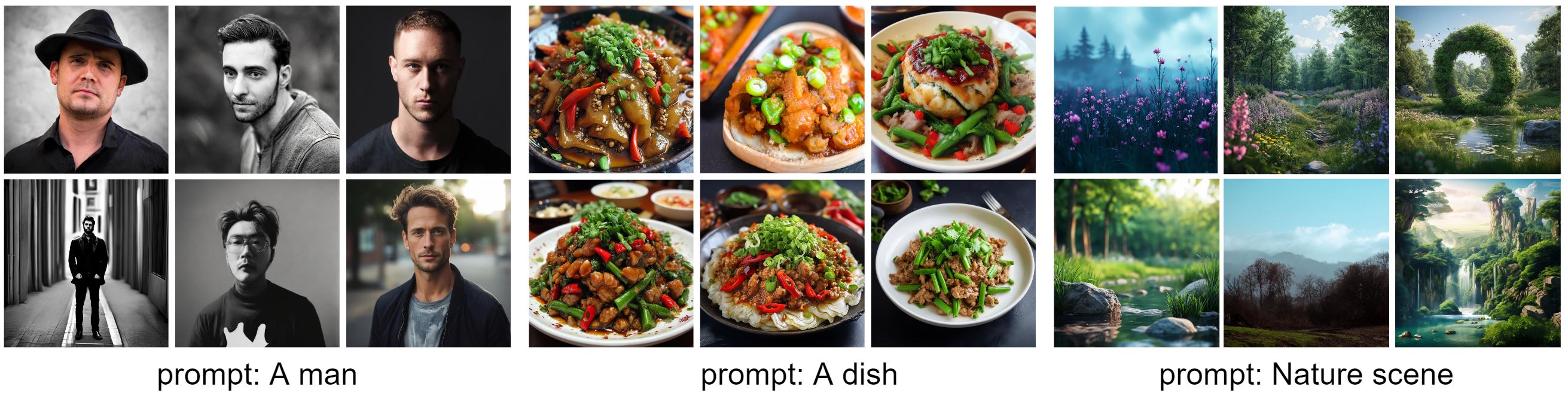}
	\caption{Image generation from PGv3 with a simple short prompt}
	\label{fig:diversity_demo}
\end{figure}

The idea behind using multi-level captions is to help the model learn a better linguistic concept hierarchy. This allows the model to not only learn the relationship between words and images but to also understand semantic relationships and hierarchies among words, by viewing the same image with varying levels of conceptual detail. As a result, our model demonstrates greater diversity when responding to short prompts involving general concepts, and adheres more closely to prompts when provided with highly detailed descriptions. Moreover, when training on datasets with fewer data samples, such as during the supervised fine-tuning (SFT) stage, multi-level captions helped prevent overfitting and significantly enhanced the model's ability to generalize the good image properties of the SFT dataset to other image domains. As shown in Fig.\ref{fig:diversity_demo}, when given a short and simple prompt, our model can generate diverse content, varying in style, color tone, ethnicity, and other attributes.

\textbf{Training stability.}
We observed loss spikes during the later stages of model training. This behavior manifested as the training loss becoming unusually large without resulting in NaN values, causing the model to enter a state where it could no longer generate meaningful content. We tried several strategies to address this issue, including lowering the betas of the Adam optimizer \cite{kingma2014adam}, using gradient clipping, and reducing the learning rate. However, none of these methods proved effective. Ultimately, we found that the following approach helped mitigate the issue.

During training, we looped through the gradients of all model parameters and counted how many gradients exceeded a specific gradient-value threshold. If this count surpassed a predefined count threshold, we discarded the training iteration, meaning we do not update the model weights using the gradients. We developed this solution after observing training behavior around the time of a loss spike. Specifically, we noticed that the number of large-valued gradients began to increase a few iterations before a spike occurred, peaking at the spike iteration. By recording the count of large-valued gradients many iterations before a spike, we set that value as a threshold to trigger the abandonment of a training iteration.

This approach differs from gradient clipping, which we found insufficient to prevent the model weights from being adversely affected by problematic gradients. Gradient clipping only reduces the magnitude of the gradients but does not alter the direction of the weight updates. As a result, the weights could still move toward collapse. By discarding the entire batch of gradients, we avoided this risk and ensured the model remained on a stable training trajectory.

\begin{figure*}
\includegraphics[width=0.9\textwidth]{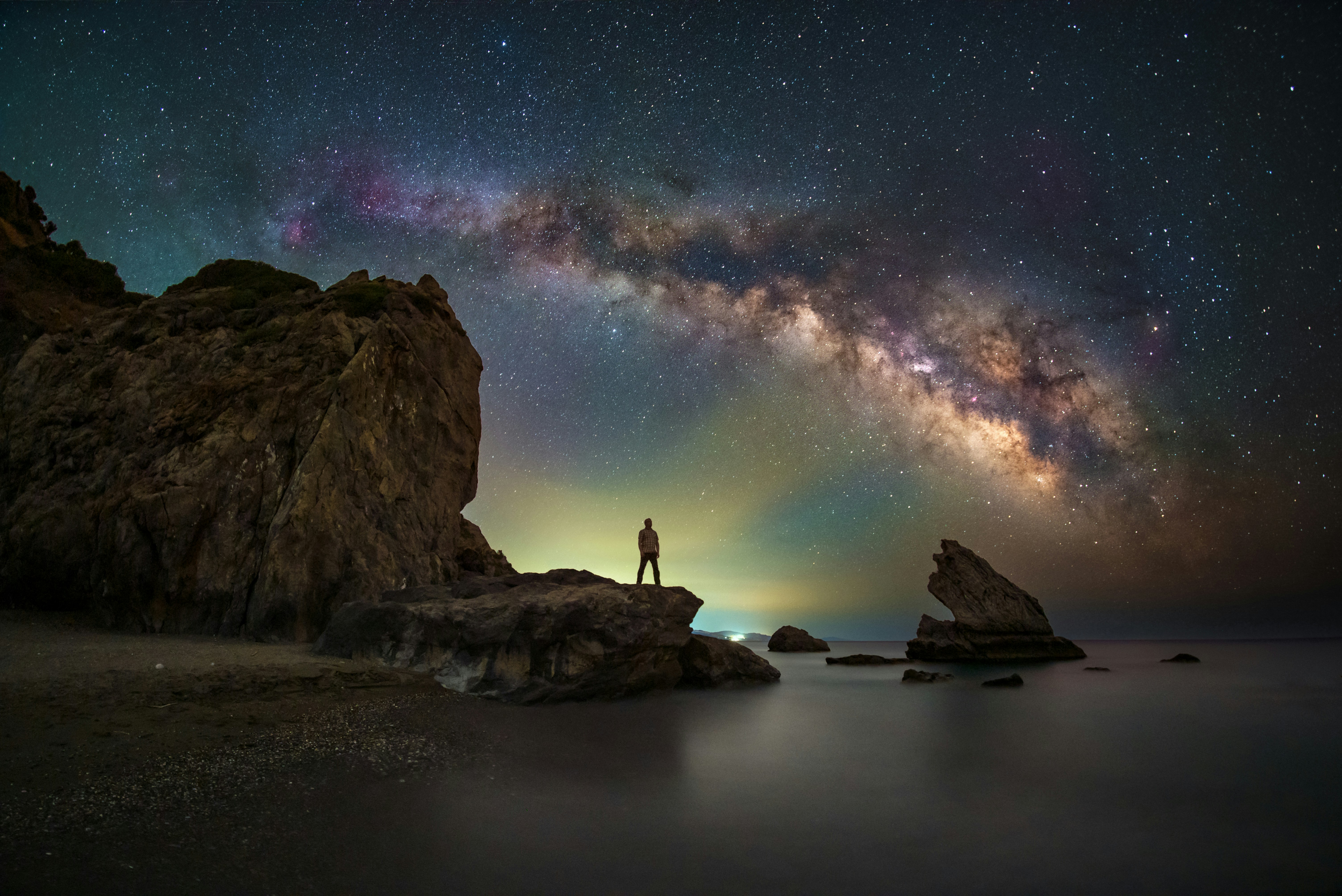}
\begin{tabular}{l}
    \multirow{8}{130mm}{\scriptsize{A breathtaking night landscape photograph captures a lone figure standing atop a jagged rock formation on a dark sandy beach with scattered pebbles, gazing up at the magnificent Milky Way galaxy spanning diagonally across the sky. The foreground features a rough-textured rock with visible striations and crevices, its weathered surface in shades of brown and gray. The midground showcases the person dressed in a checkered shirt and pants, silhouetted against the soft glow of ambient light from an unseen horizon. To the left stands a massive, rugged cliff with uneven surfaces, visible cracks, and patches of green vegetation clinging to its slopes. On the right side of the midground, another distinctive rock structure extends out into the calm sea, partially submerged and silhouetted against the starry backdrop. The background is dominated by the Milky Way's vibrant display of colors, including deep navy blue, emerald green, lavender purple, burnt orange, and white hues. A subtle aurora-like effect in pale yellow and green adds depth to the scene below the stars. The water appears smooth and glassy due to a long exposure technique, creating a serene atmosphere with gentle ripples reflecting the faint light above. Small rocks are scattered across the surface near the shore. Distant lights hint at nearby human activity along the coastline. The overall mood is awe-inspiring and tranquil, enhanced by the high contrast between the dark elements and the brightly illuminated stars, creating a harmonious blend of natural beauty and cosmic wonder.}}
\end{tabular}
\vspace{30mm}

\caption{Qualitative example of our in-house captioner.}
\label{fig:captioner}

\end{figure*}
\section{Captioning Model}
\label{sec:captioner}
\subsection{PG Captioner}

For the captioning model, we used a standard vision-language architecture that consists of a vision encoder, a vision-language adapter, and 
a decoder-only language model \cite{liu2024improved, li2023blip, chen2023pali}. Following recent works \cite{beyer2024paligemma}, we confirmed that a high-resolution vision encoder is very important for detailed captioning of the images. We also found an inverse relationship between the model size and the data volume required to train it to a reasonable quality: small models require much more data to produce coherent long captions compared to larger models. For model training, we used synthetic data generation workflows \cite{sharifzadeh2024synth, hong2024cogvlm2, xiao2024florence} and iterative self-improvement \cite{fang2024vila}.

\subsection{CapsBench}

Captioning evaluation is a complex problem. There are two main classes of captioning evaluation metrics. The first one is reference-based metrics - BLEU \cite{papineni2002bleu}, CIDEr \cite{vedantam2015cider}, METEOR \cite{banerjee2005meteor}, SPICE \cite{anderson2016spice}. These metrics use a ground truth caption or a set of captions to calculate similarity as a quality metric. The problem with such methods is that the score is bound to the reference format. Recent works, like FAIEr \cite{wang2021faier} propose a combination of visual and textual scene graphs to create better metrics. The second class of evaluation metrics is reference-free metrics - CLIPScore \cite{hessel2022clipscore}, InfoMetIC \cite{hu2023infometic}, TIGEr \cite{jiang2019tiger}. These methods use semantic vectors from the reference image or multiple regions of the image to calculate a similarity metric for the proposed captions. The drawback of these methods is that for dense images and long and detailed captions, semantic vectors will be not be representative as they will include too many concepts. 

A new type of evaluation method is a question-based metric. The authors of QACE \cite{lee2021qace} proposed a framework for generating questions from captions and evaluating proposed captions using these questions. A similar approach was recently proposed for evaluation of image generation models \cite{Cho2024DSG}. Davidsonian Scene Graph (DSG) allows generation of questions organized in dependency graphs that help to comprehensively evaluate text-to-image models.

Inspired by DSG and DPG-bench \cite{hu2024ella}, we propose a reversed approach for image captioning evaluation. Based on the image, we generate "yes-no" question-answer pairs across 17 categories: general, image type, text, color, position, relation, relative position, entity, entity size, entity shape, count, emotion, blur, image artifacts, proper noun (world knowledge), color palette, and color grading. The answer to the majority of questions is "yes", as this provides a more clear signal of correctness. During evaluation, we use an LLM to answer the questions based solely on the candidate caption. Available answer options are "yes", "no", and "n/a" (when it is impossible to answer the question). We then compare the answers with the reference answers to calculate the resulting accuracy.

We are releasing CapsBench\footnote{\url{https://huggingface.co/datasets/playgroundai/CapsBench}} , a set of 200 images and 2471 questions for them, resulting in 12 questions per image on average. Images represent a wide variety of types: film scenes, cartoon scenes, movie posters, invitations, advertisements, casual photography, street photography, landscape photography, and interior photography. A diversity of questions and images enables comprehensive evaluation of image captioning systems. We describe evaluation results for PG Captioner and other state-of-the-art models in Section \ref{sec:caption_eval}.
\section{Image qualitative results}
\label{sec:qualitative}
\subsection{Photo-realism}
\begin{figure}[H]
	\centering
	\includegraphics[width=13cm,height=6.5cm]{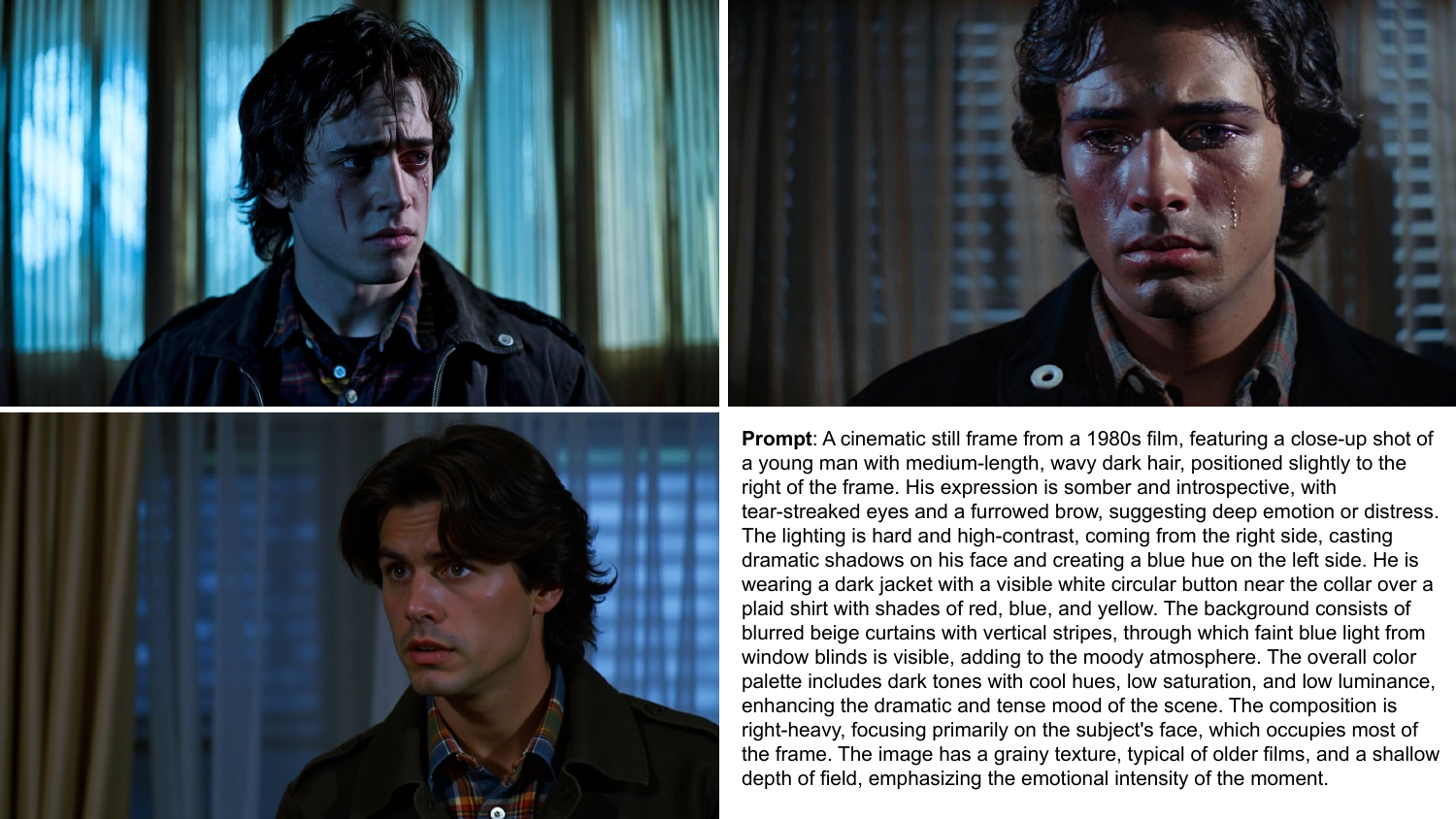}
        \includegraphics[width=13cm,height=6.5cm]{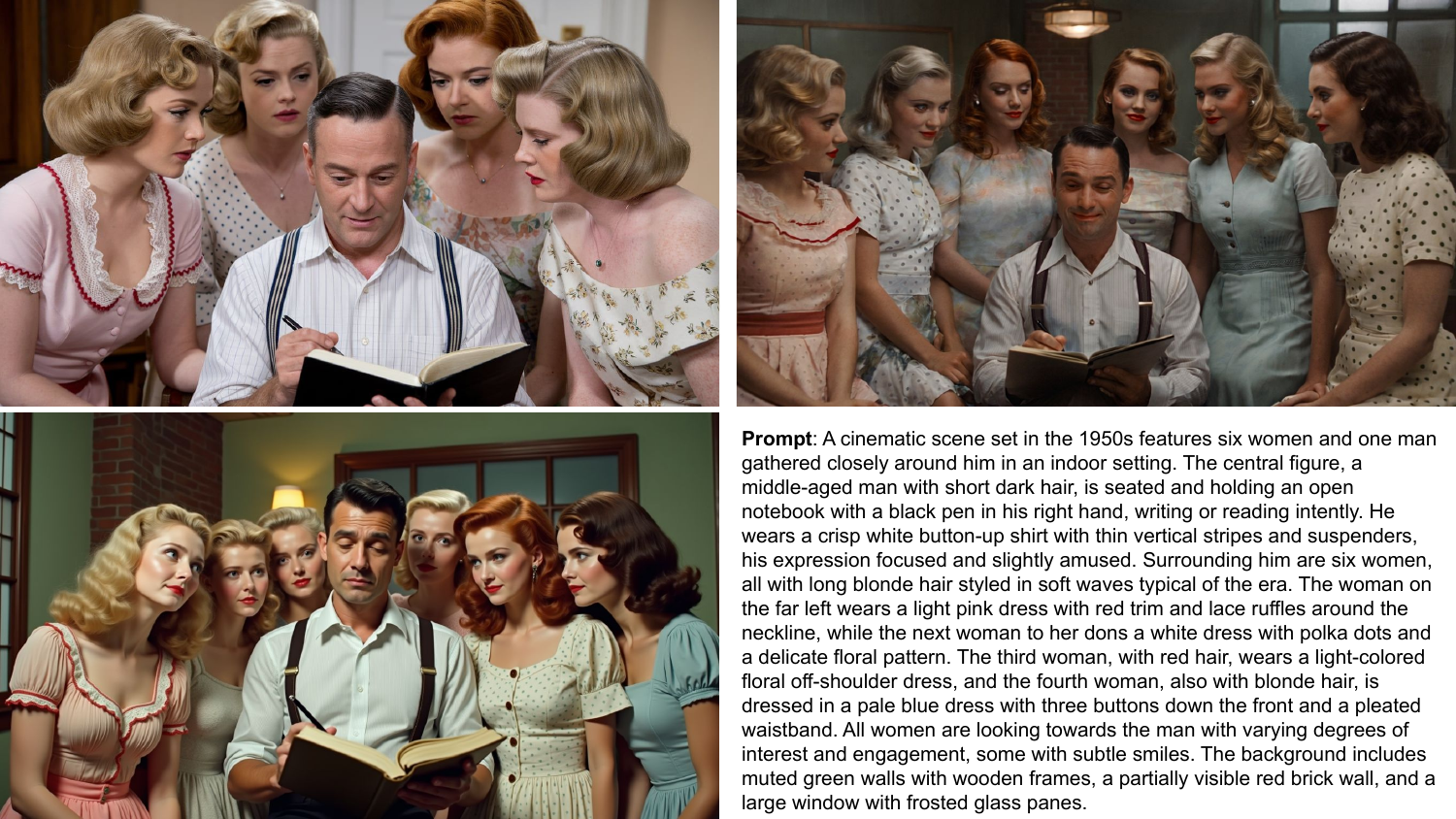}
        \includegraphics[width=13cm,height=6.5cm]{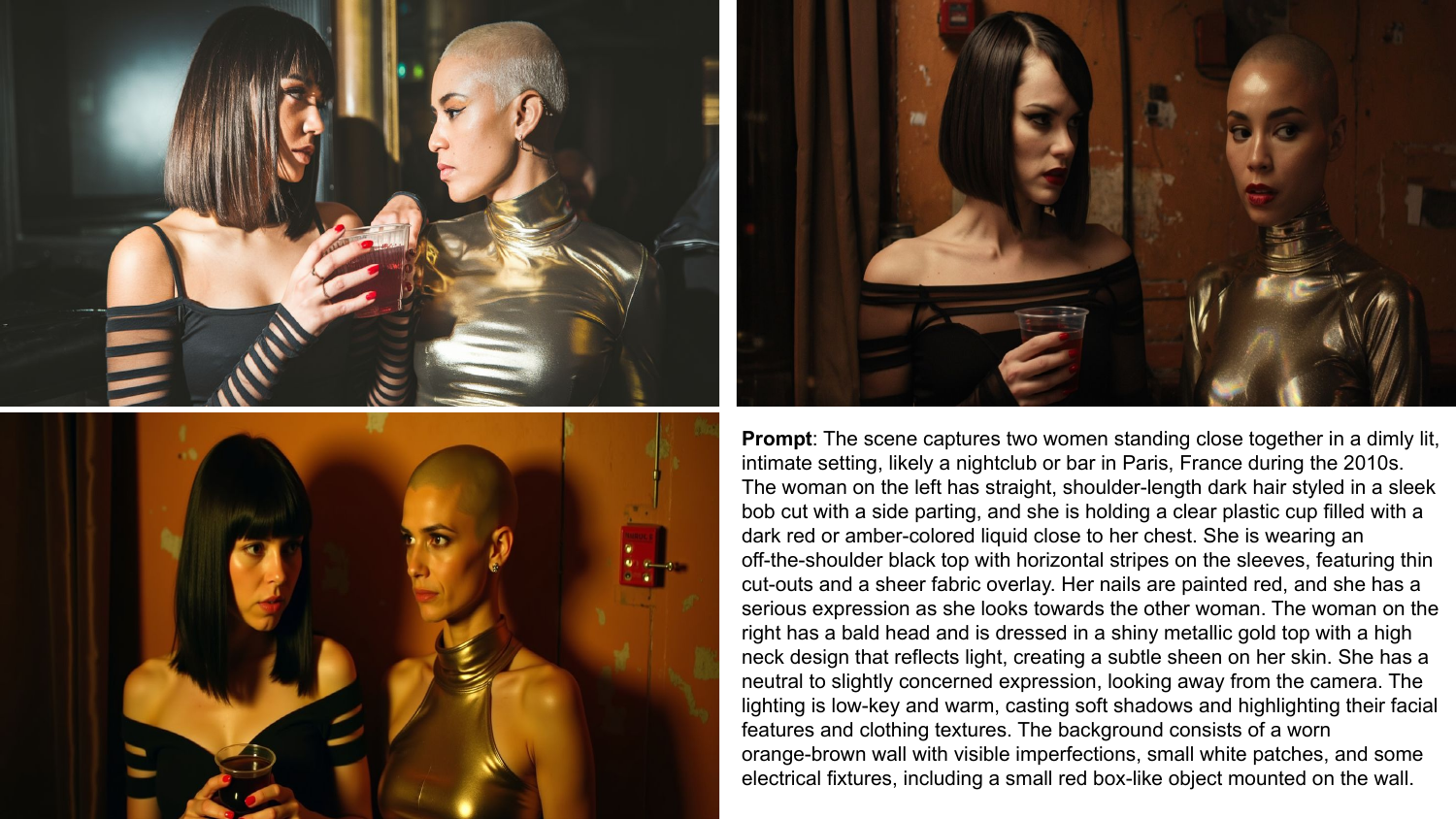}
	\caption{Photo-realistic qualitative comparison:  the top-left is Ideogram-2, the top-right is PGv3, the
bottom-left is Flux-pro, and the bottom-right is the prompt. Zoom in for better comparison on details and textures.}
	\label{fig:photoreal_demo_1}
\end{figure}

\begin{figure}[t]
	\centering
        \includegraphics[width=13.5cm]{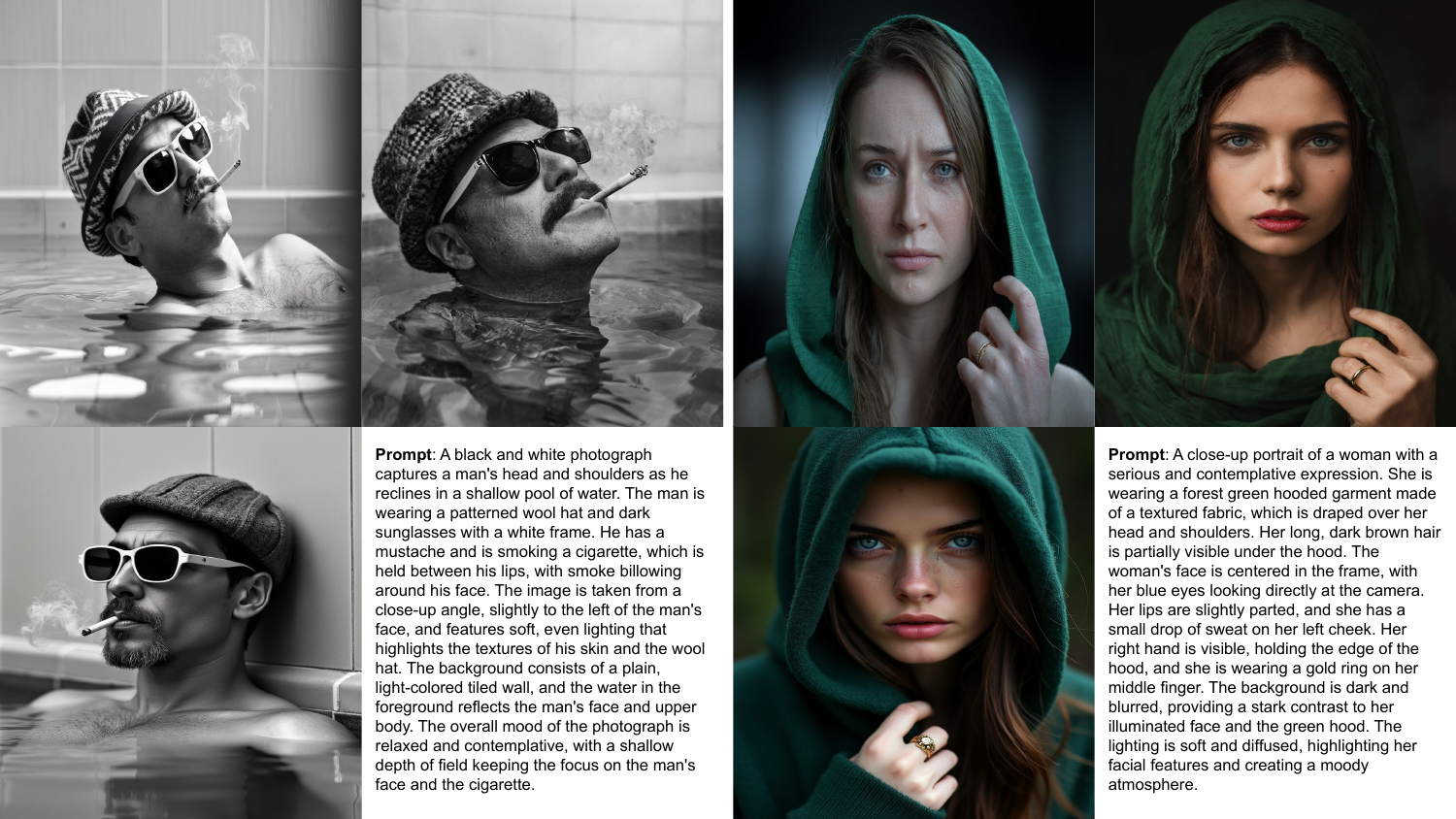}
        \caption{Photo-realistic qualitative comparison cont.:  the top-left is Ideogram-2, the top-right is PGv3, the
bottom-left is Flux-pro, and the bottom-right is the prompt. Zoom in for better comparison.}
	\label{fig:photoreal_demo_2}
\end{figure}

\begin{figure}[t]
	\centering
        \includegraphics[width=13.5cm]{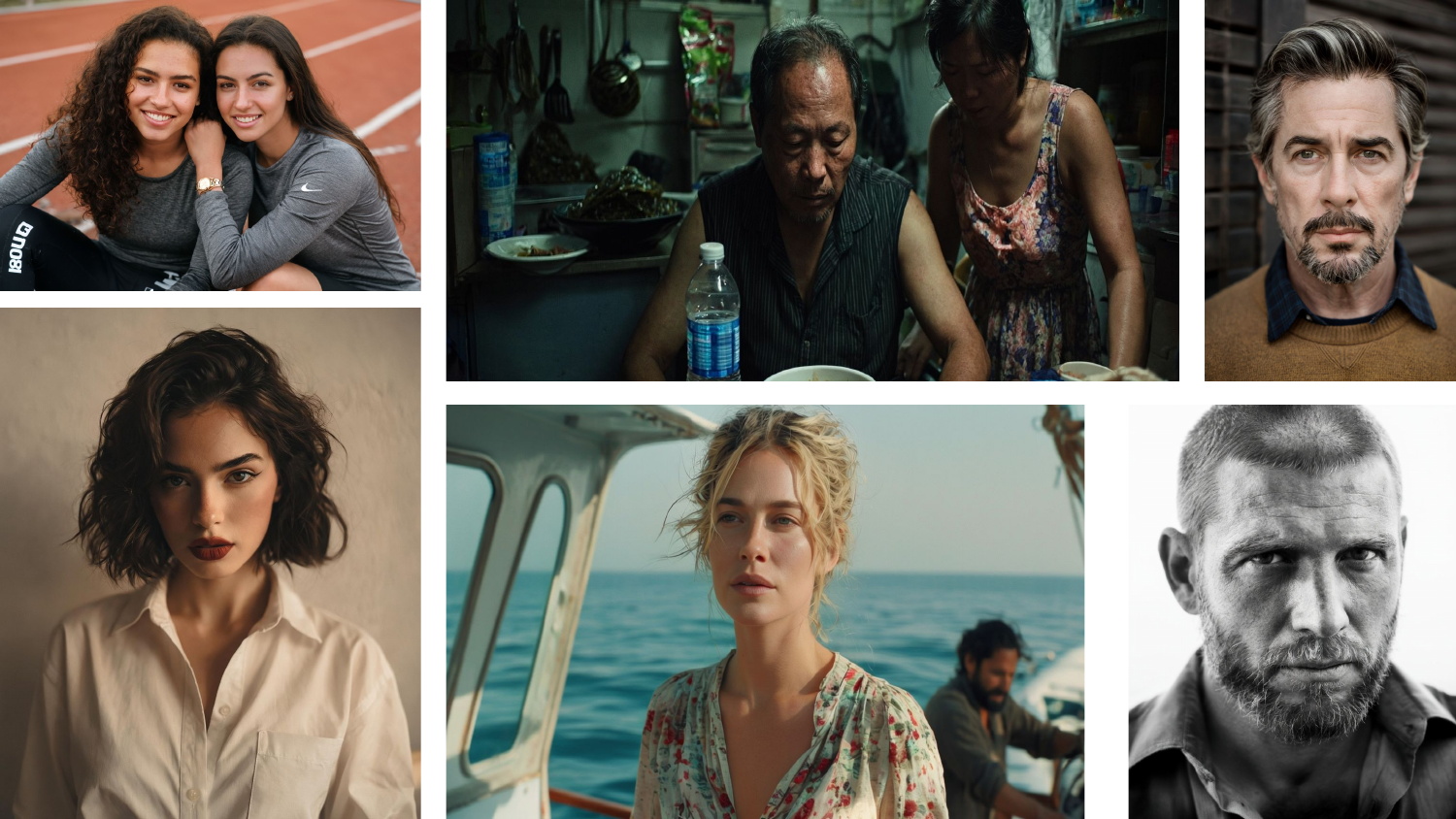}
        \caption{Photo-realistic qualitative results from PGv3. Non-curated random samples from our evaluation prompt set.}
	\label{fig:photoreal_demo_3}
\end{figure}

Fig.~\ref{fig:photoreal_demo_1}, \ref{fig:photoreal_demo_2}, and \ref{fig:photoreal_demo_3} show image generation results from PGv3, alongside other top-tier models. In each panel from Fig.~\ref{fig:photoreal_demo_1} and \ref{fig:photoreal_demo_2}, the images are arranged as follows: the top-left is Ideogram-2, the top-right is PGv3, the bottom-left is Flux-pro, and the bottom-right is the prompt. When viewed as thumbnails, the images from all 3 models appear similar with minimal qualitative differences. 

However, upon zooming in to examine details and textures, notable distinctions become apparent. Flux-pro consistently produces images with overly smooth skin textures, resembling oily or 3D renderings, rather than realistic depictions. Ideogram-2 offers more realistic skin textures compared to Flux-pro but struggles with prompt adherence, often missing key details as prompts grow longer. In contrast, PGv3 excels in both prompt-following and generating realistic images. Furthermore, PGv3 demonstrates a significantly better cinematic quality than the other models.

\subsection{Prompt-Following}
\begin{figure}[H]
	\centering
        \includegraphics[width=13cm,height=5cm]{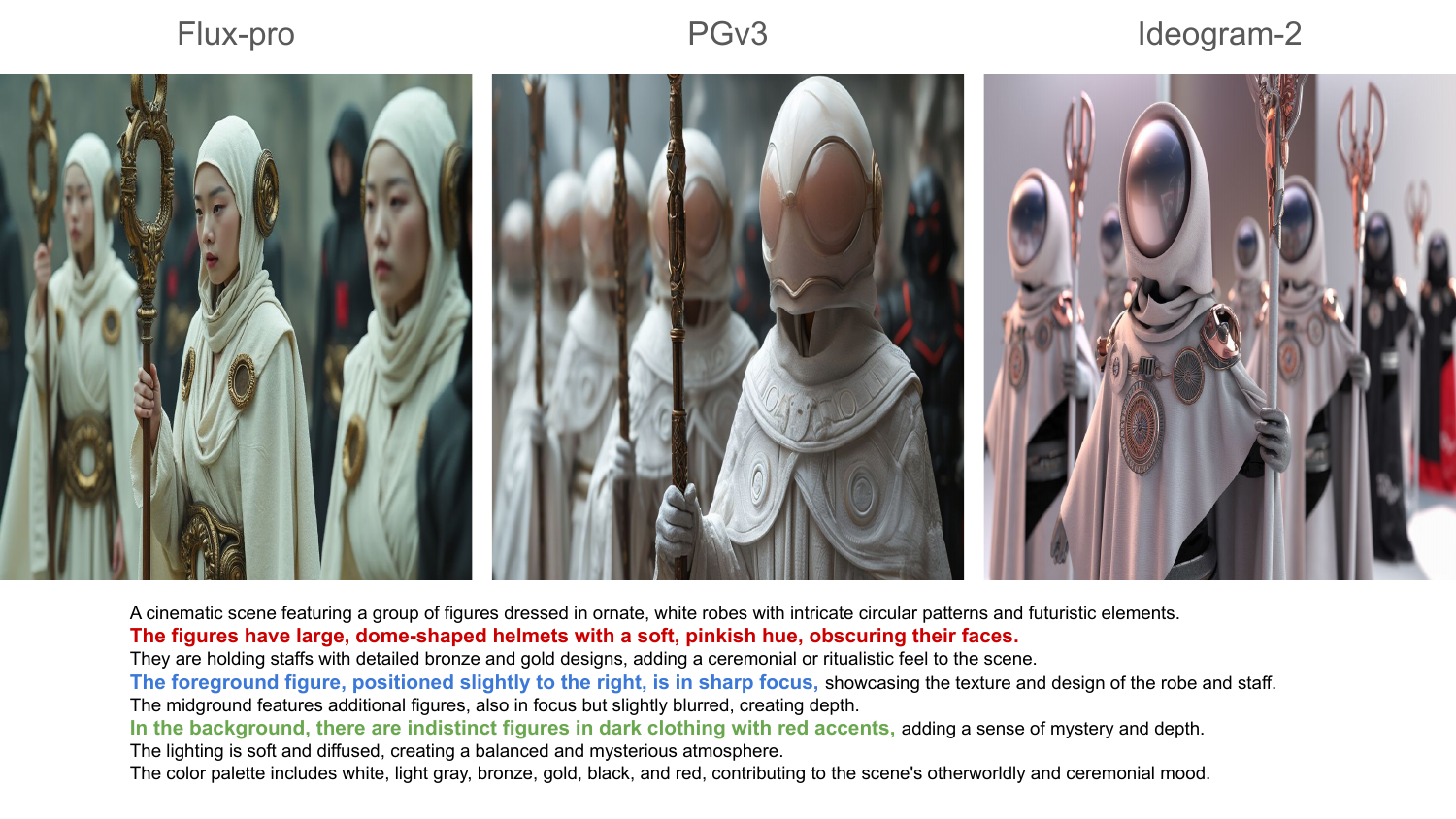}
        \includegraphics[width=13cm,height=6cm]{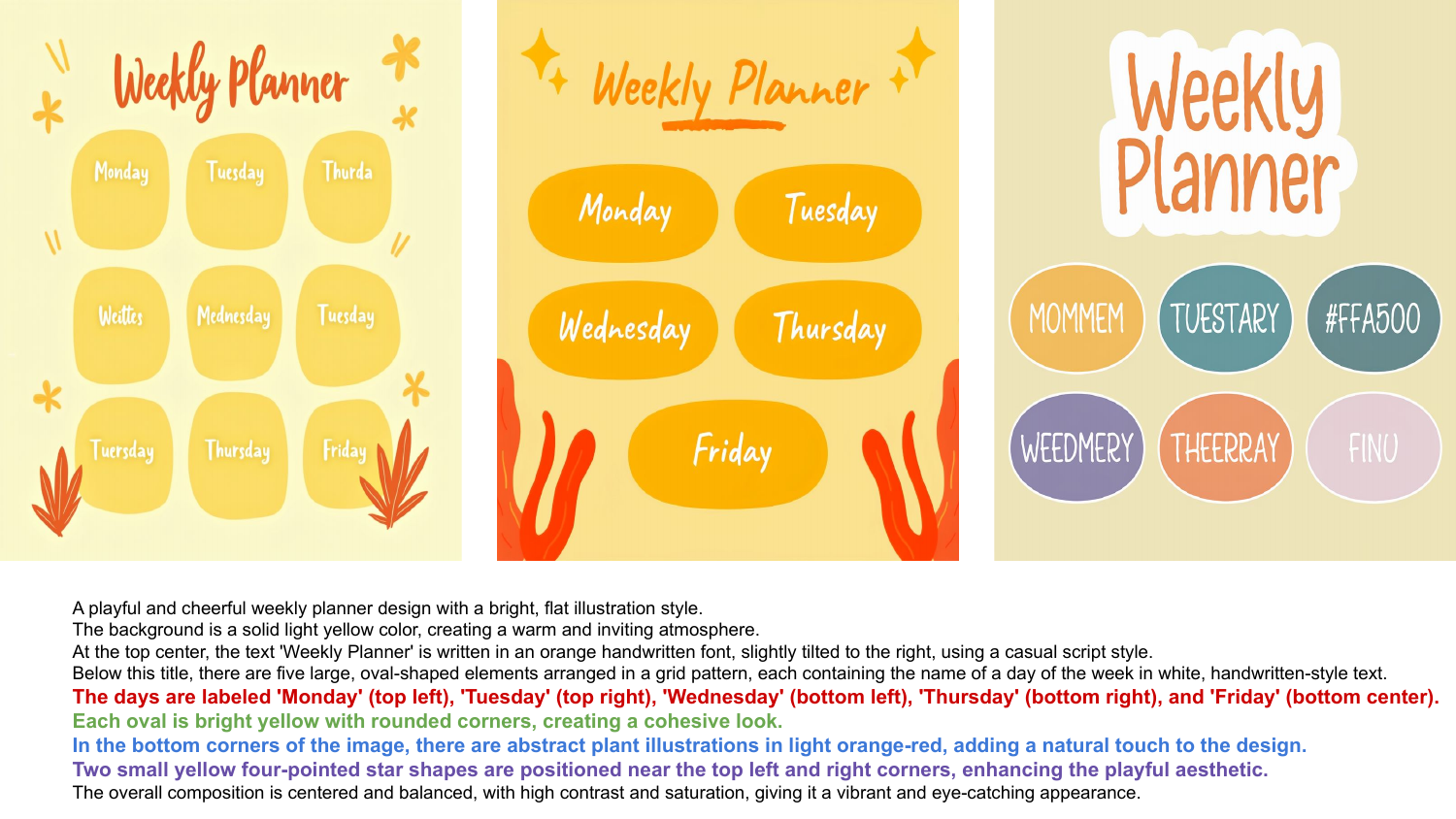}
        \includegraphics[width=13cm,height=7cm]{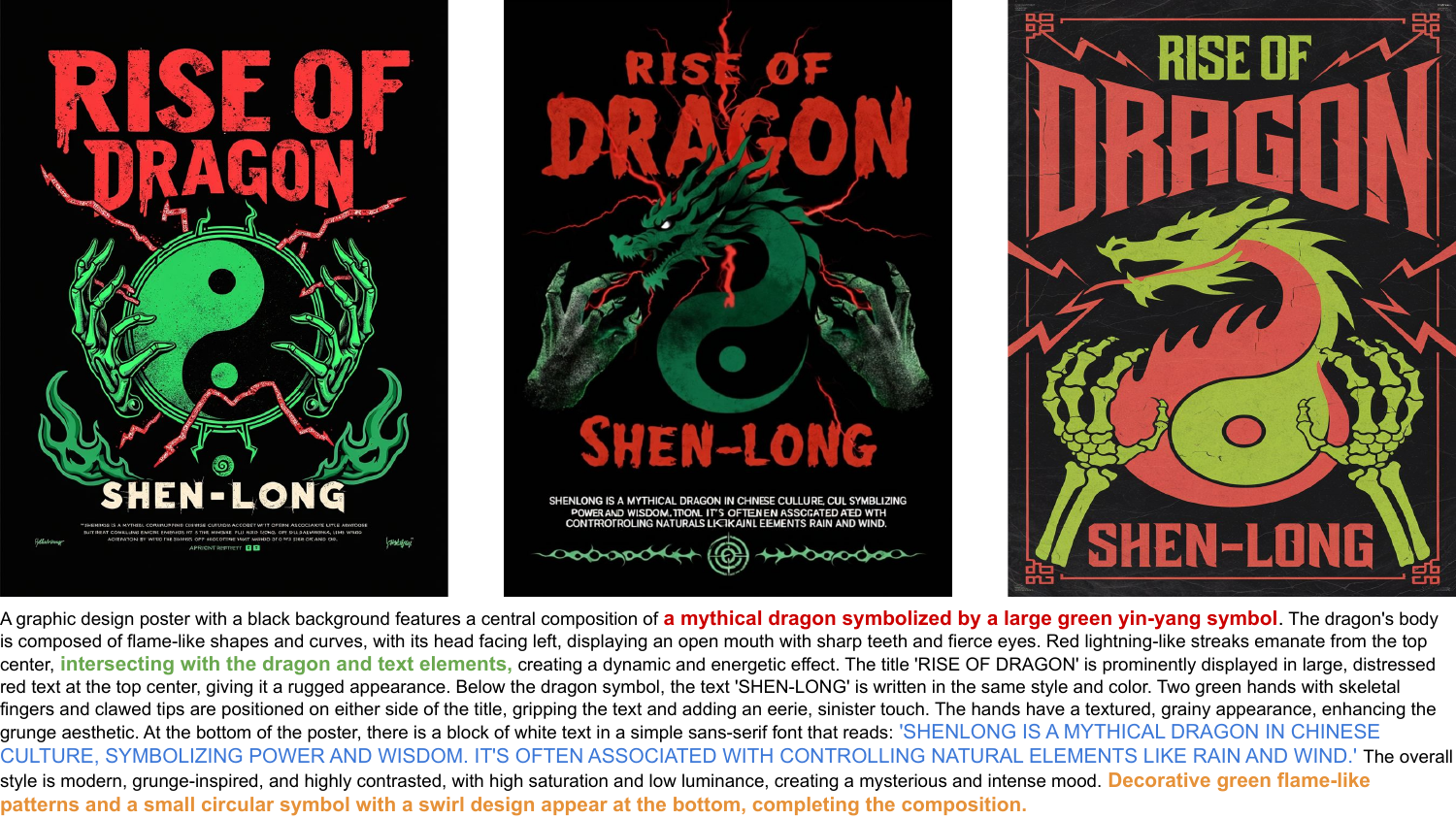}
        \caption{Prompt-following qualitative comparison. Highlighted text in bright colors indicate instances where Flux-pro or Ideogram-2 failed to follow the prompts, while PGv3 consistently adheres to all of the details in the prompt. The examples shown are selected samples from our evaluation prompt set.}
	\label{fig:prompt_follow_demo_1}
\end{figure}

Fig.\ref{fig:prompt_follow_demo_1} illustrates three examples highlighting the superior prompt-following ability of our PGv3 model compared to other models, including Ideogram-2 and Flux-pro. In each prompt, the colored text highlights the specific details that other models fail to capture, while PGv3 consistently follows those details. There is no particular meaning to specific colors. The advantage of PGv3 becomes particularly evident as the testing prompts become longer and contain more detailed information. We attribute this performance to our LLM-integrated model structure and our advanced VLM captioning system.

\subsection{Text Rendering}
\begin{figure}[t]
	\centering
        \includegraphics[width=13.5cm]{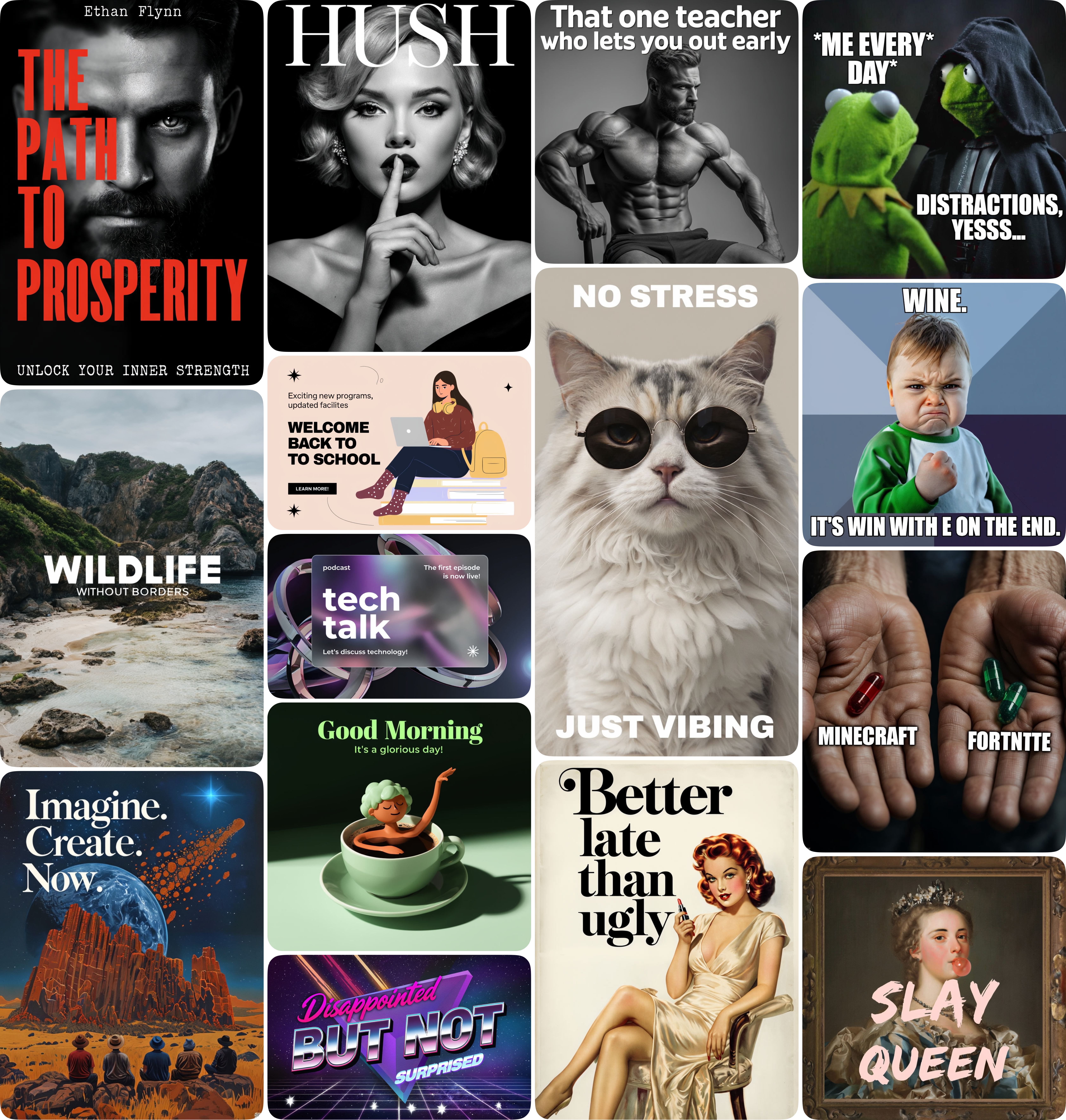}
        \caption{Text rendering qualitative results. PGv3 can generate text-rich content across various categories, ranging from professional designs like advertisements and logos to playful creations such as memes and greeting cards.}
	\label{fig:text_demo_1}
\end{figure}

Fig.\ref{fig:text_demo_1} demonstrates the text rendering capabilities of PGv3. The model can generate images across a wide range of design categories, including posters, logos, stickers, book covers, and presentation slides. Notably, PGv3 can reproduce trending memes with customized text and create entirely new memes with limitless characters and compositions, thanks to its powerful prompt-following and text rendering abilities.

\clearpage

\subsection{RGB Color Control}

\begin{figure}[H]
	\centering
        \includegraphics[height=8cm,width=12cm]{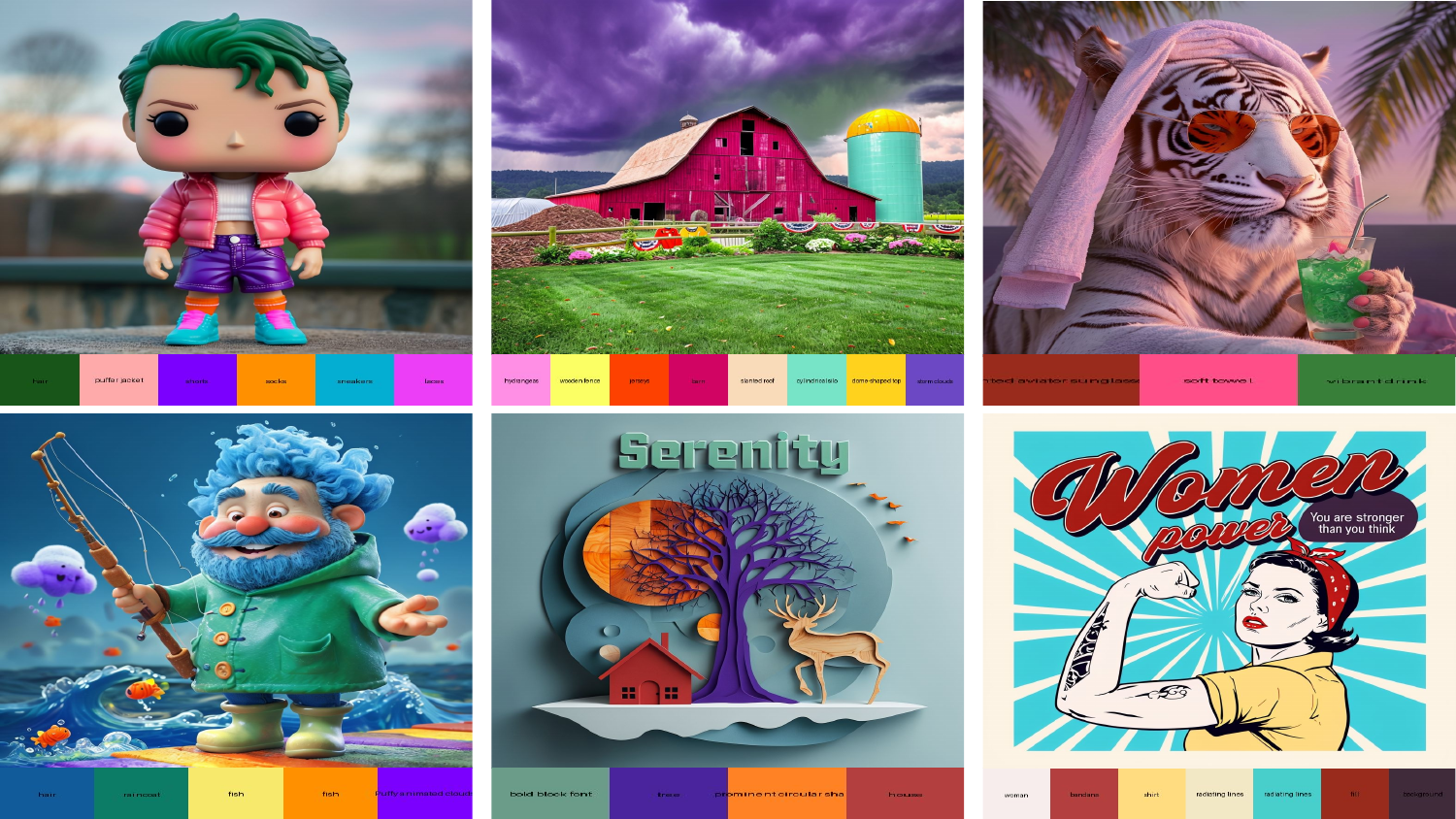}
        \caption{RGB color control qualitative results. Prompts are omitted due to space limitation, the color bar under each image indicates the specified items and colors in the prompt}
	\label{fig:rgb_demo_1}
\end{figure}

\begin{figure}[H]
	\centering
        \includegraphics[width=12cm]{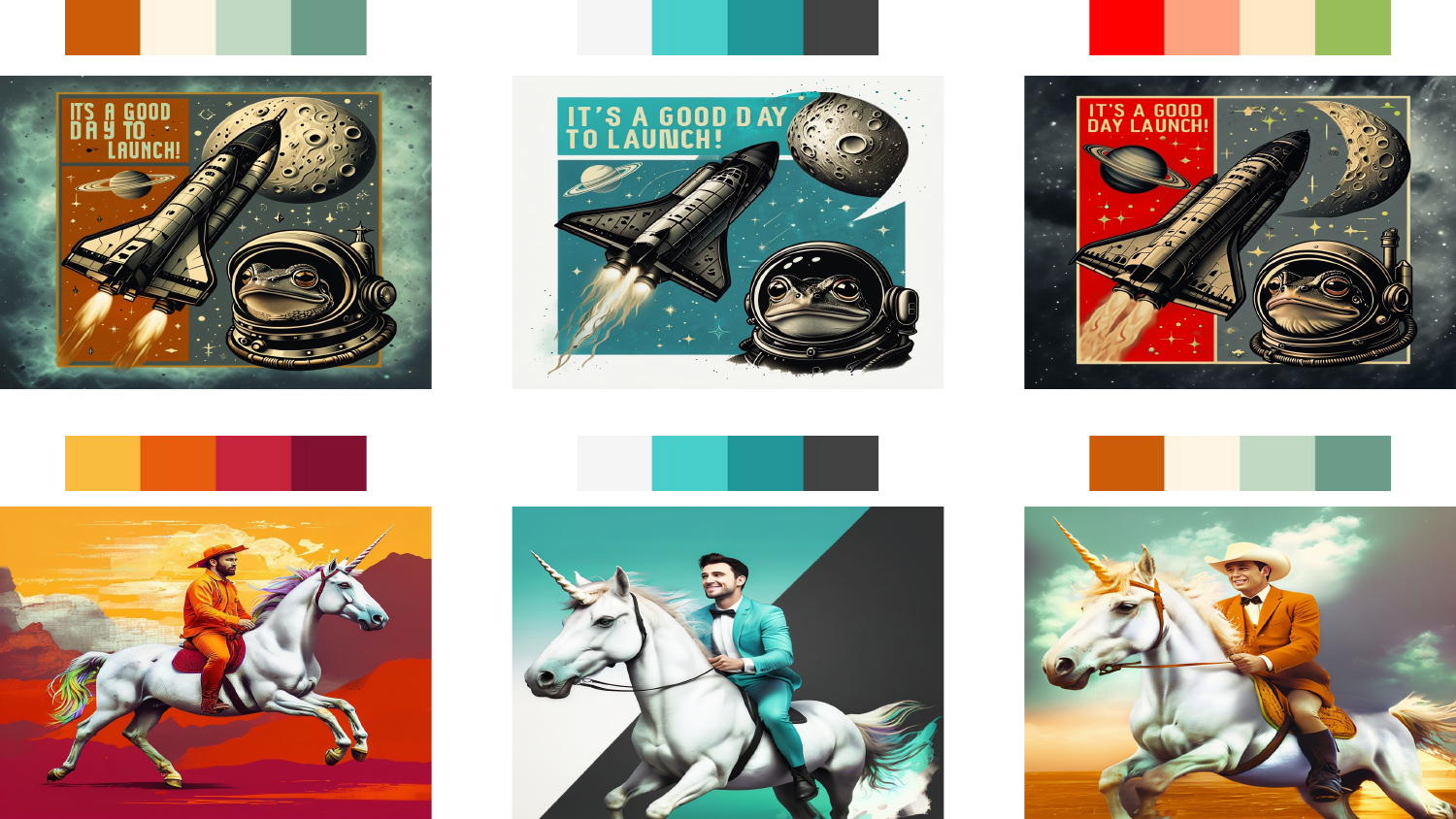}
        \caption{RGB color palette control qualitative results. PGv3 accepts an overall color palettes, automatically applying the specified colors to appropriate objects and areas.}
	\label{fig:rgb_demo_2}
\end{figure}

PGv3 achieves exceptionally fine-grained color control in generated content, surpassing standard color palettes. With its robust prompt-following capabilities and specialized training, PGv3 enables users to precisely control the color of each object or area within an image using exact RGB values, making it ideal for professional design scenarios that require precise color matching. As shown in Fig.~\ref{fig:rgb_demo_1} and \ref{fig:rgb_demo_2}, PGv3 accepts both overall color palettes, automatically applying the specified colors to appropriate objects and areas, and dedicated color values for specific objects as defined by the user. 


\clearpage

\subsection{Multilingual Prompt Input}

\begin{figure}[H]
	\centering
        \includegraphics[width=13.5cm,height=6cm]{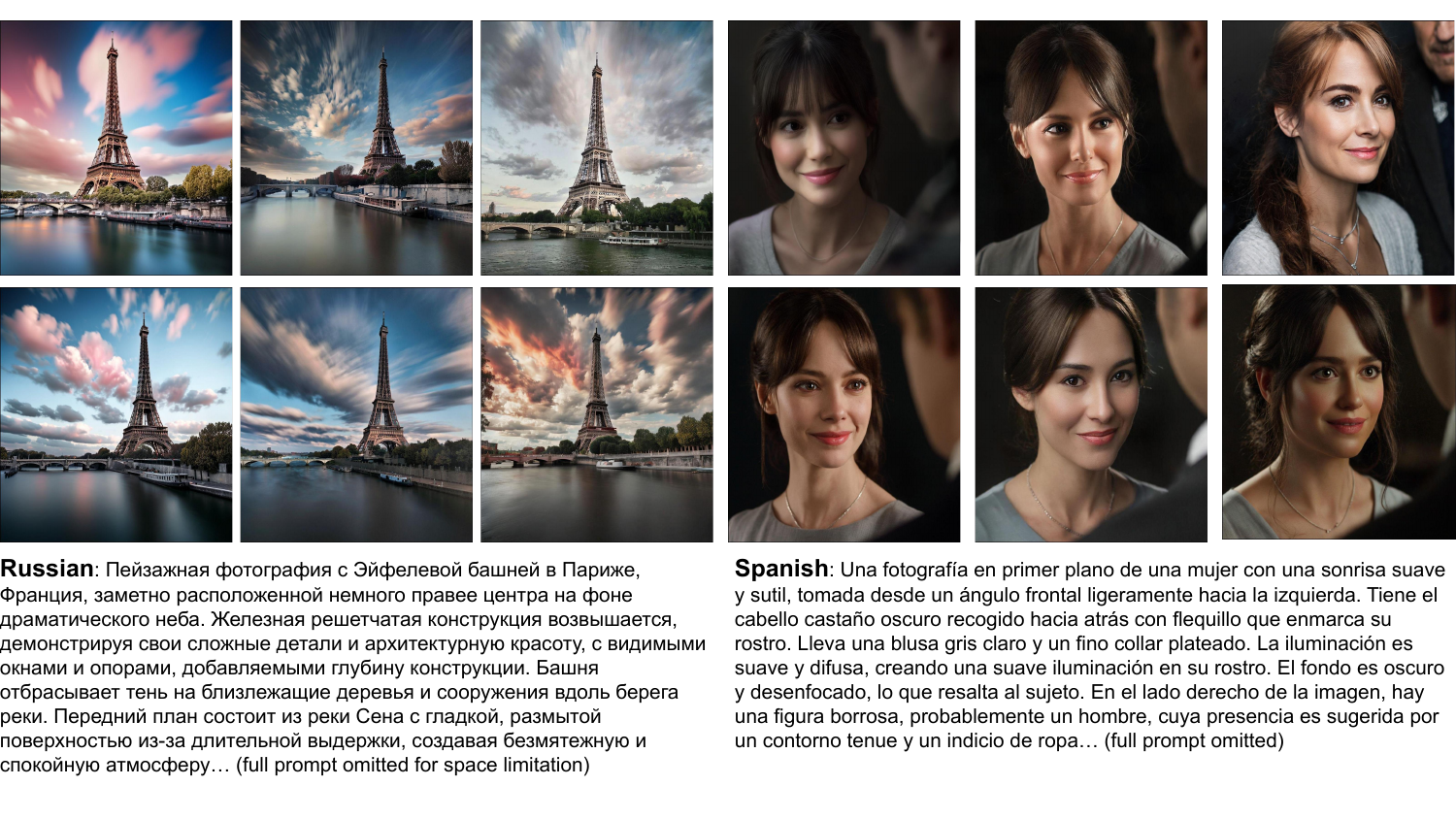}
        \includegraphics[width=13.5cm,height=6cm]{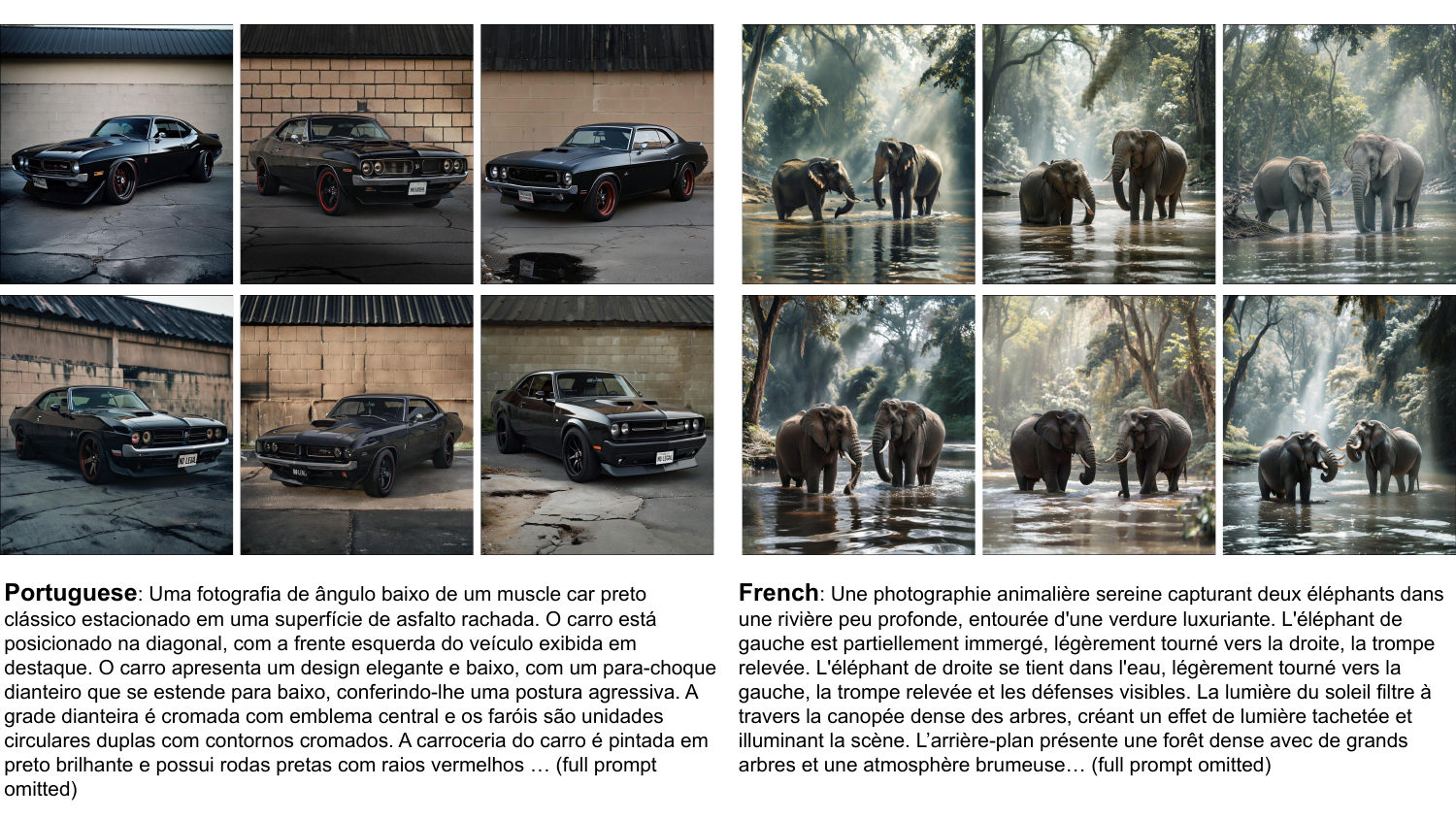}
        \includegraphics[width=13.5cm,height=6cm]{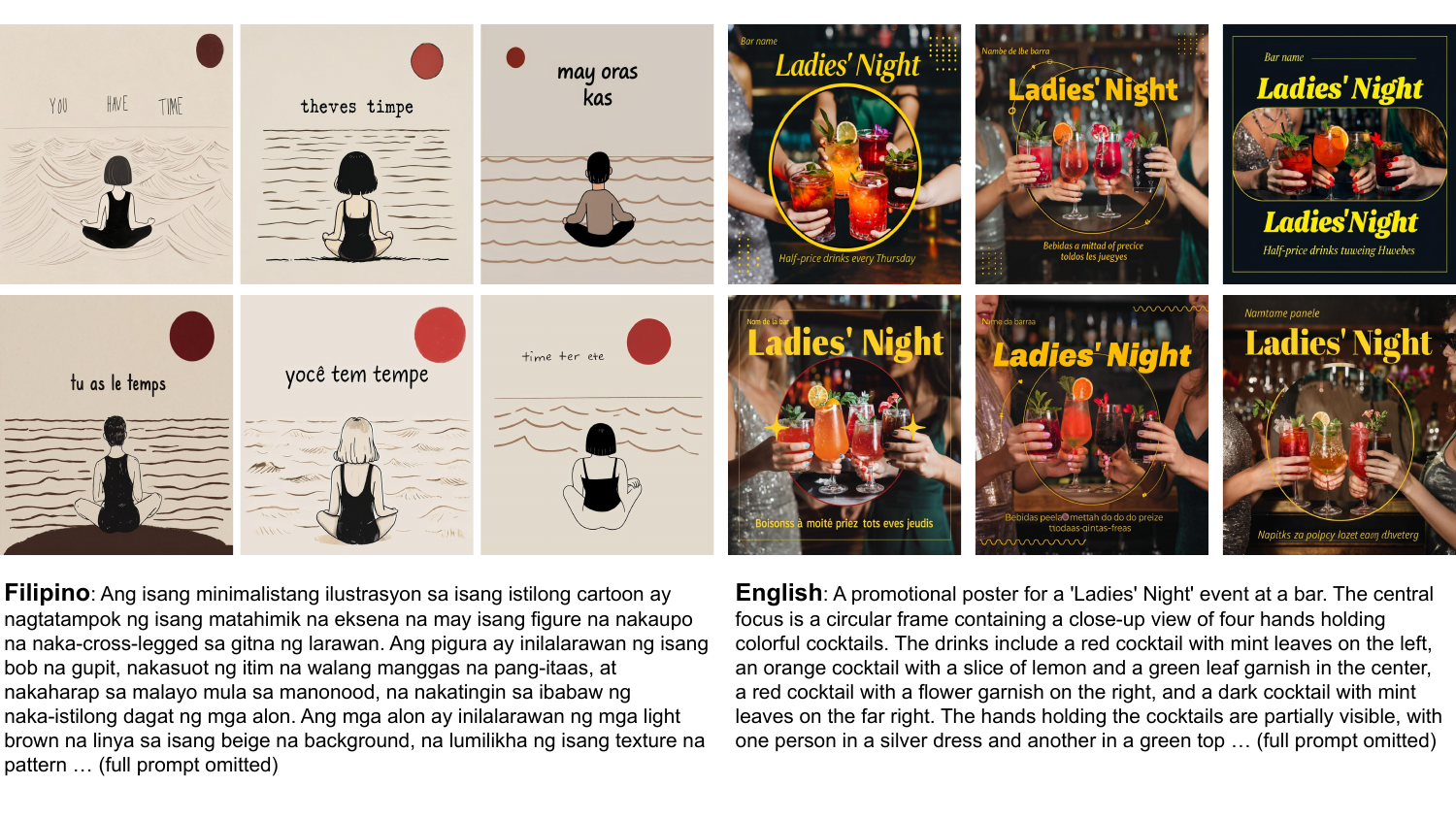}
        \caption{Multilingual qualitative results. In each panel, images are generated from prompts in English, Spanish, Filipino, French, Portuguese, and Russian, arranged from top left to bottom right. For each panel, we display the prompt in one of the languages used, and all languages are represented across the panels.}
	\label{fig:multilingual_demo_1}
\end{figure}

Thanks to the powerful LLM used as our text encoder, which inherently understands multiple languages and forms well-related word representations across them, PGv3 can naturally interpret prompts in various languages. We found that this multilingual capability is fully realized with just a small dataset of multilingual text-and-image pairs—tens of thousands of images are sufficient.

Fig.~\ref{fig:multilingual_demo_1} illustrates the model's performance across various languages. In each panel, images are generated from prompts in English, Spanish, Filipino, French, Portuguese, and Russian, arranged from top left to bottom right. PGv3 consistently produces content that adheres to the prompts in each respective language. It is important to note that we did not train our model on non-English text rendering data; therefore, PGv3 typically does not recognize or generate text in other languages.
\section{Image Model Quantitative Evaluation}
\label{sec: image-eval}
\subsection{Evaluation on Graphic Design Ability}
\begin{figure*}[t]
\centering
\vspace{-5mm}
\includegraphics[width=0.9\textwidth]{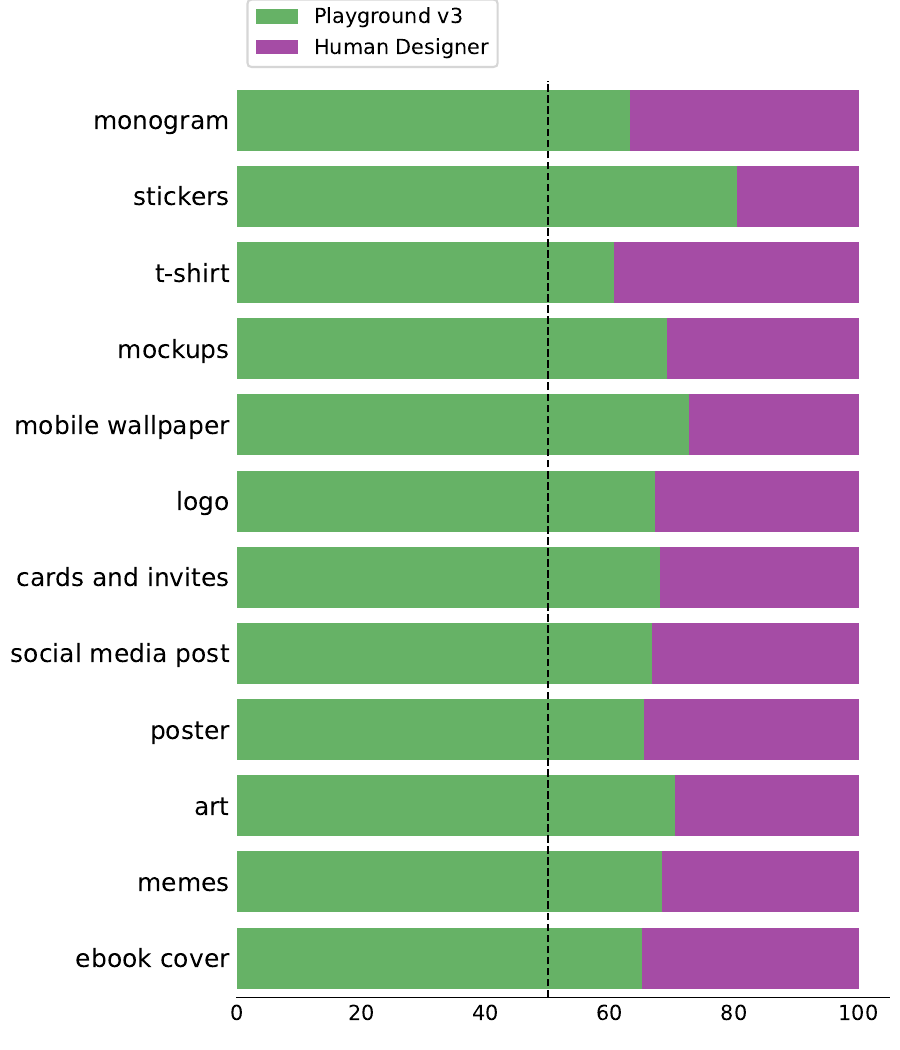}
\caption{Graphic design ability evaluation. We performed user preference studies on common use cases requiring graphic design skills. We compared our model Playground v3 and high-quality ground truth data created by designers that can be used to represent the average graphic design ability of humans. From this study, users preferred designs generated by our model in all categories, especially for stickers, art, and mobile wallpaper.
}
\label{fig:ability}
\end{figure*}
We collected around 4k high-quality images created by human designers in common graphic design applications like \textit{monograms, stickers, t-shirt, mockups, mobile wallpaper, logo, cards, posters} etc. We captioned the images referred as Human in Fig.~\ref{fig:ability} with our in-house captioner, PG Captioner. We then used the same text caption to generate images using Playground v3. We conducted user preference studies on the paired data and asked users to select which image is preferred. Each pair got at least 7 votes, and we used majority voting to determine the winning image.

\subsection{Image-text Alignment Evaluation}
\subsubsection{DPG-bench}

\begin{table*}[t]
\vspace{-2mm}
\centering
\setlength{\tabcolsep}{6pt}
\renewcommand{\arraystretch}{1.2}
\caption{DPG-bench evaluation. We compared our method with SoTA open-source models. From the table, our model achieved high text-image alignment performance.}
\begin{tabular}{lcccccc}
\toprule 
    Method & Overvall & Global & Entity & Attribute & Relation & Other \\ 
    \midrule
    SDXL~\cite{podell2023sdxl} & 74.65 & 83.27 & 82.43 & 80.91 & 86.76 & 80.41 \\
    Playground v2.5~\cite{li2024playground} & 75.47 & 83.06 & 82.59 & 81.20 & 84.08 & 83.50 \\
    Lumina-Next~\cite{zhuo2024luminanextmakingluminat2xstronger} & 74.63 & 82.82 & 88.65 & 86.44 & 80.53 & 81.82 \\
    Hunyuan-DiT~\cite{li2024hunyuan} & 78.87 & 84.59 & 80.59 & 88.01 & 74.36 & 86.41 \\
    PixArt-Sigma~\cite{chen2024pixartsigmaweaktostrongtrainingdiffusion} & 80.54 & 86.89 & 82.89 & 88.94 & 86.59 & 87.68 \\
    DALLE 3~\cite{betker2023improving} & 83.50 & 90.97 & 89.61 & 88.39 & 90.58 & 89.83 \\
    SD3-Medium~\cite{esser2024scalingrectifiedflowtransformers} & 84.08 & 87.90 & 91.01 & 88.83 & 80.70 & 88.68 \\
    Playground v3 (ours) & \textbf{87.04} & \textbf{91.94} & 85.71 & \textbf{90.90} & 90.00 & \textbf{92.72} \\

\bottomrule
\end{tabular}
\label{tbl:dpg}
\end{table*}

In order to access the text following ability (text-image consistency), we compared our models with SoTA models on two benchmarks. We measured DPG-bench follows \cite{hu2024ella} in Tab.~\ref{tbl:dpg}, and our model achieved an overall score of 87.04, which is higher than DALLE 3 and SD3-medium.

\subsubsection{DPG-bench Hard}
However, upon closer inspection of the results, we found that the Visual-Question-Answering (VQA) model used in the open-sourced DPG-bench often makes errors, providing incorrect answers even when models correctly generate the content specified by the testing prompts. To more accurately assess the true prompt-following performance, we developed an upgraded version called "DPG-bench Hard." The following outlines the steps used to construct this benchmark.

First, we applied k-means clustering on our internally curated dataset, which consists of images rated by real users. This resulted in 120 clusters covering diverse image domains. From each cluster, we randomly selected 20 images near the cluster centers, yielding a testing set of 2,400 images in total. We used GPT-4o to caption these images, ensuring the captions were detailed and comprehensive, with an average length of about 256 tokens. To avoid potential prompt-style bias favoring our model, we particularly choose to not use our in-house Captioner for this task. After generating the captions, we asked GPT-4o to create yes-or-no questions based on each caption, then categorized these questions in line with the original DPG-bench. Each prompt generated an average of 30 questions. This process parallels how we constructed our \textbf{CapsBench}, which will be described in detail in  Sec.\ref{sec:caption_eval}

To evaluate each model, we generated four images per testing prompt using random seeds, then used GPT-4o to answer the corresponding questions for each image. We summarized the "yes" responses from GPT-4o as the final accuracy rate. Despite the possibility of occasional errors by GPT-4o, we found the error margin to be below 5\%, making the testing scores more closely aligned with human judgment as compared to the original DPG-bench.

\begin{figure*}[t]
\centering
\vspace{-5mm}
\includegraphics[width=0.9\textwidth]{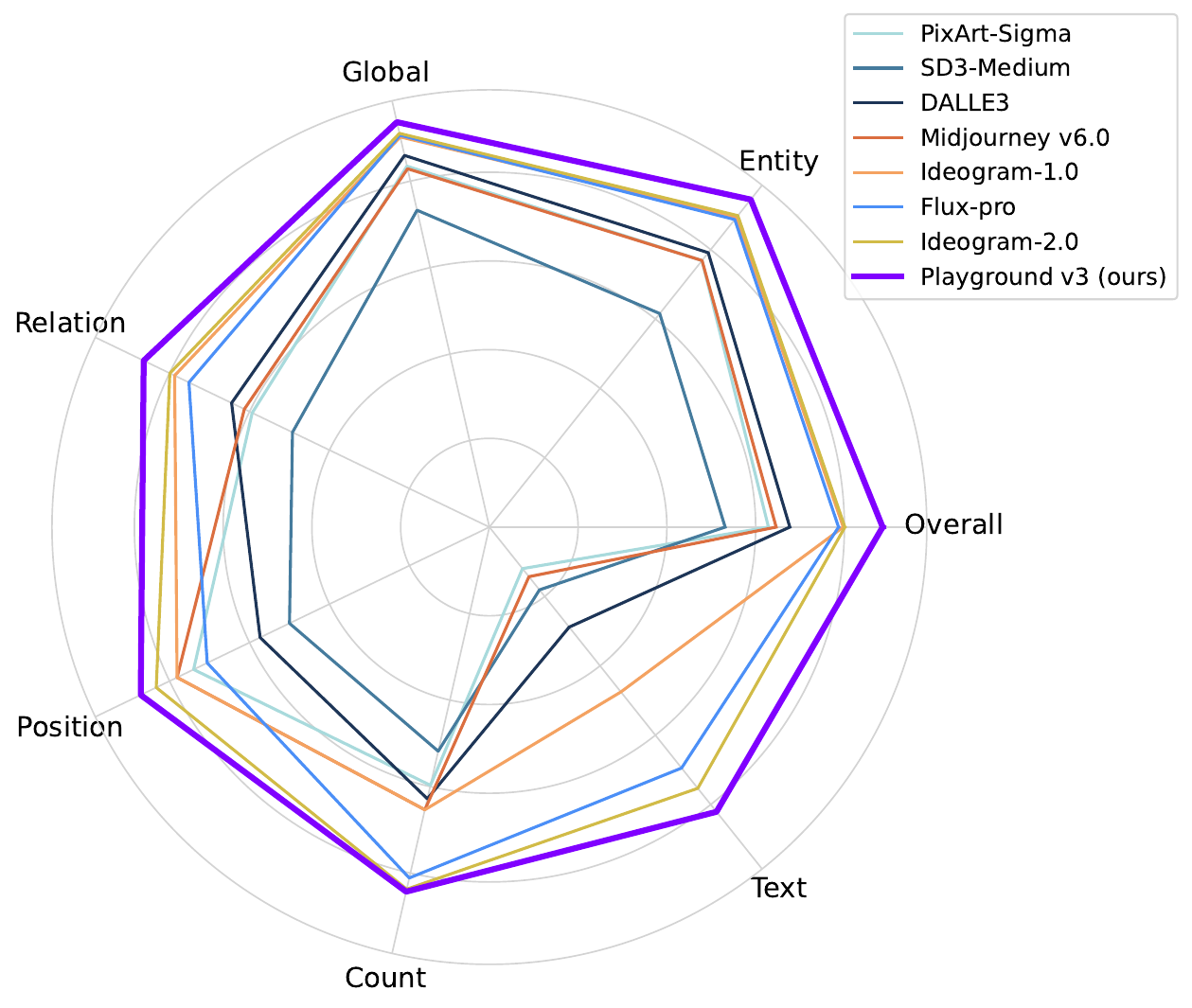}
\caption{Visual presentation of DPG-bench Hard, our internal benchmark powered by GPT4-o}
\label{fig:dpg}
\end{figure*}

Fig.\ref{fig:dpg} and Tab.~\ref{tbl:dpg-hard} present our testing results. PGv3 achieved an overall accuracy score of 88.62\%, outperforming other tested models, including Flux-pro and Ideogram-2, across all sub-categories. Notably, a significant portion (40\%) of the testing images in our DPG-bench Hard set belong to graphic design-related categories, such as posters, ads, greeting cards, and book covers, all of which require the model to be able to follow the complicated object arrangements and layout requirements from the testing prompts. This benchmark not only reflects PGv3's leading prompt-following performance but also effectively highlights PGv3’s superior performance in graphical design domains.

\begin{figure}[t]
	\centering
        \includegraphics[width=13.5cm,height=7.5cm]{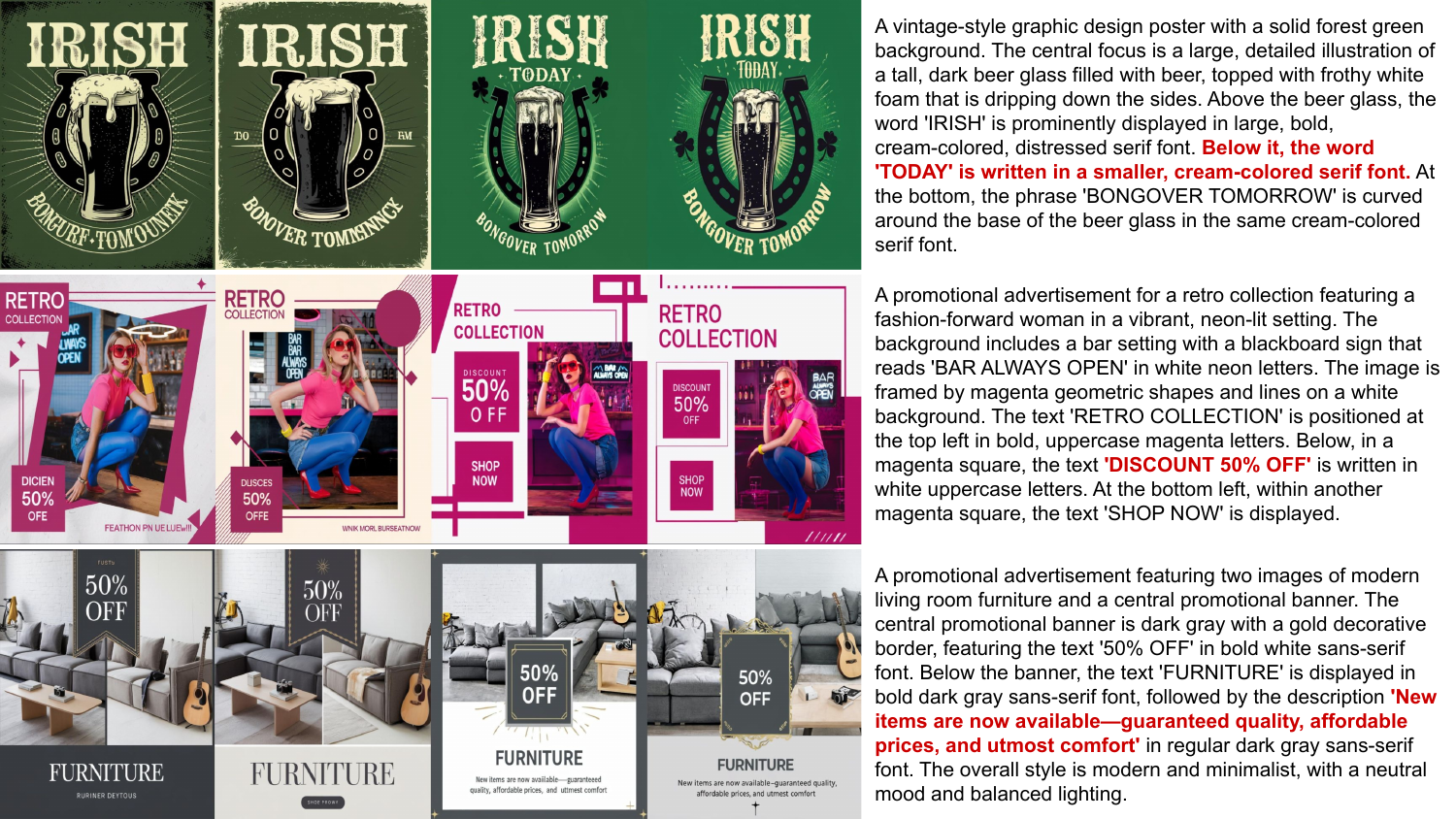}
        \includegraphics[width=13.5cm,height=7.5cm]{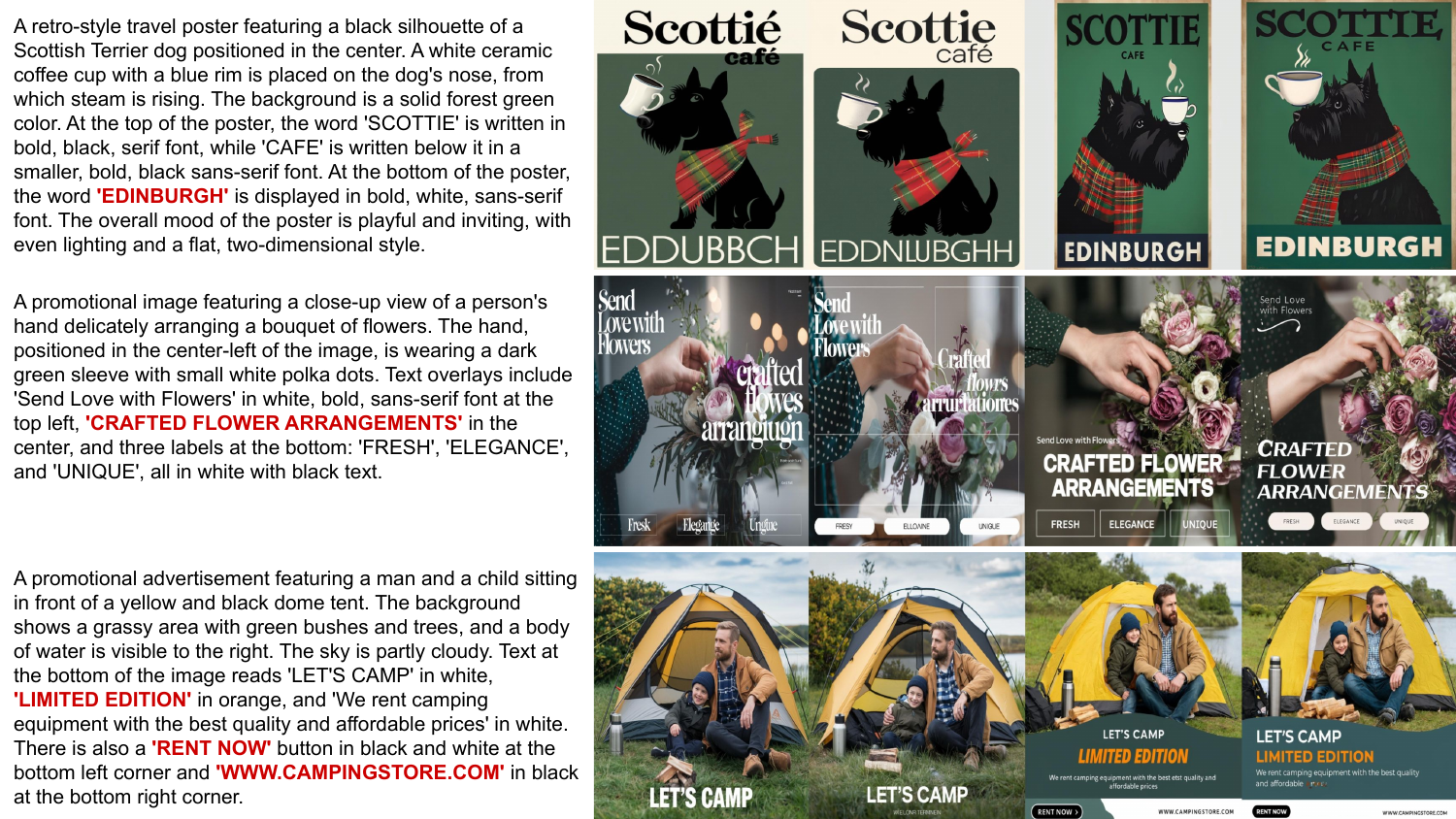}
        \caption{Text rendering comparison between Ideogram-2 and PGv3 using testing prompts from our DPG-bench Hard. In each panel, the left two images are random samples from Ideogram-2, while the right two images are random samples from PGv3. Due to space constraints, testing prompts are abbreviated. Text highlighted in bold red indicates areas where Ideogram-2 makes errors, while PGv3 performs accurately.}
	\label{fig:text_compare_ideogram}
\end{figure}

On text synthesis accuracy, PGv3 scores 82\%, outperforming Flux-pro's 69\% and Ideogram-2's 75\%. A closer analysis reveals that, besides common mistakes like misspellings and corrupted glyph generation, another primary reason behind the lower scores of Flux-pro and Ideogram-2 is their tendency to omit required text elements as the prompts increase in length and complexity. This issue highlights PGv3's superior prompt-following ability, which ensures more accurate text rendering even with longer, more detailed prompts. Examples on text rendering image results from PGv3 together with Ideogram-2 can be found in Fig.\ref{fig:text_compare_ideogram}. While it primarily shows failed cases from Ideogram-2, we want to emphasize that Ideogram-2 generally performs well, and both models have good and bad testing cases. Tab.~\ref{tbl:dpg-hard} provides the non-cherry-picked quantitative performance scores for these models.

We believe that our DPG-bench Hard offers a comprehensive evaluation of text-to-image model performance. As such, we focused our efforts on collecting images from Flux-pro and Ideogram-2 specifically for DPG-bench Hard. For subsequent evaluations in this section, we did not collect additional images from these models, as the extensive image collection process requires significant effort, which we deemed unnecessary.

\subsection{Image-Text Reasoning Evaluation}
We evaluated the model's reasoning ability using a popular benchmark, GenEval \cite{ghosh2023genevalobjectfocusedframeworkevaluating}. In Tab~\ref{tbl:geneval}, our model achieved a 0.76 overall score, higher than SD3's 0.74. In particular, our model has a much better score on object and position reasoning.

\begin{table*}[t]
\vspace{-2mm}
\centering
\setlength{\tabcolsep}{6pt}
\renewcommand{\arraystretch}{1.2}
\caption{Quantitative results from DPG-bench Hard, our internal benchmark powered by GPT4-o}
\begin{tabular}{lccccccc}
\toprule 
    Method & Overvall & Entity & Global & Relation & Position & Count & Text \\ 
    \midrule
    SD3-Medium~\cite{esser2024scalingrectifiedflowtransformers} & 53.13 & 61.57 & 73.27 & 49.25 & 50.00 & 51.82 & 18.10 \\
    PixArt-Sigma~\cite{chen2024pixartsigmaweaktostrongtrainingdiffusion} & 62.89 & 76.85 & 83.46 & 59.33 & 73.98 & 59.71 & 11.97 \\
    Midjourney v6.0 & 64.63 & 76.86 & 82.86 & 61.29 & 78.12 & 65.36 & 14.30 \\
    DALLE 3~\cite{betker2023improving} & 67.71 & 79.11 & 85.94 & 64.48 & 57.35 & 62.82 & 28.84 \\
    Flux-pro~\cite{Flux} & 78.69 & 88.68 & 90.62 & 75.15 & 70.58 & 81.13 & 69.47 \\
    Ideogram-1.0~\cite{Ideogram} & 79.75 & 89.32 & 90.21 & 78.73 & 78.12 & 65.36 & 47.53\\
    Ideogram-2.0~\cite{Ideogram} & 80.12 & 89.81  & 91.04 & 79.93 & 83.33 &  83.72  & 75.32 \\
    Playground v3 (ours) & \textbf{88.62} & \textbf{94.45} & \textbf{93.64} & \textbf{86.47} & \textbf{87.21} & \textbf{84.30} & \textbf{82.09} \\

\bottomrule
\end{tabular}
\label{tbl:dpg-hard}
\end{table*}

\begin{table*}[h]
\vspace{-2mm}
\centering
\setlength{\tabcolsep}{6pt}
\renewcommand{\arraystretch}{1.2}
\caption{GenEval evaluation benchmark. We measured Flux-dev by using their public model API, since there are no official numbers reported. For the rest of the baselines, we used their official numbers in the paper.}
\begin{tabular}{lccccccc}
\toprule 
    Method & Overall & Single & Two & Counting & Colors & Position & Attribution \\ 
    \midrule
    SDXL~\cite{podell2023sdxl} & 0.55 & 0.98 & 0.74 & 0.39 & 0.85 & 0.15 & 0.23 \\
    DALLE 3~\cite{betker2023improving} & 0.67 & 0.96 & 0.87 & 0.47 & 0.83 & 0.43 & 0.45 \\
    SD3~\cite{esser2024scalingrectifiedflowtransformers} & 0.74 & 0.99 & 0.94 & 0.72 & 0.89 & 0.33 & 0.60 \\
    *Flux-dev~\cite{Flux} & 0.68 & 0.99 & 0.85 & 0.74 & 0.79 & 0.21 & 0.48 \\
    Playground v3 & \textbf{0.76} & 0.99 & \textbf{0.95} & 0.72 & 0.82 & \textbf{0.50} & 0.54 \\

\bottomrule
\end{tabular}
\label{tbl:geneval}
\end{table*}

\subsection{Text Synthesis Evaluation}
\begin{figure*}[t]
\includegraphics[width=1.0\textwidth]{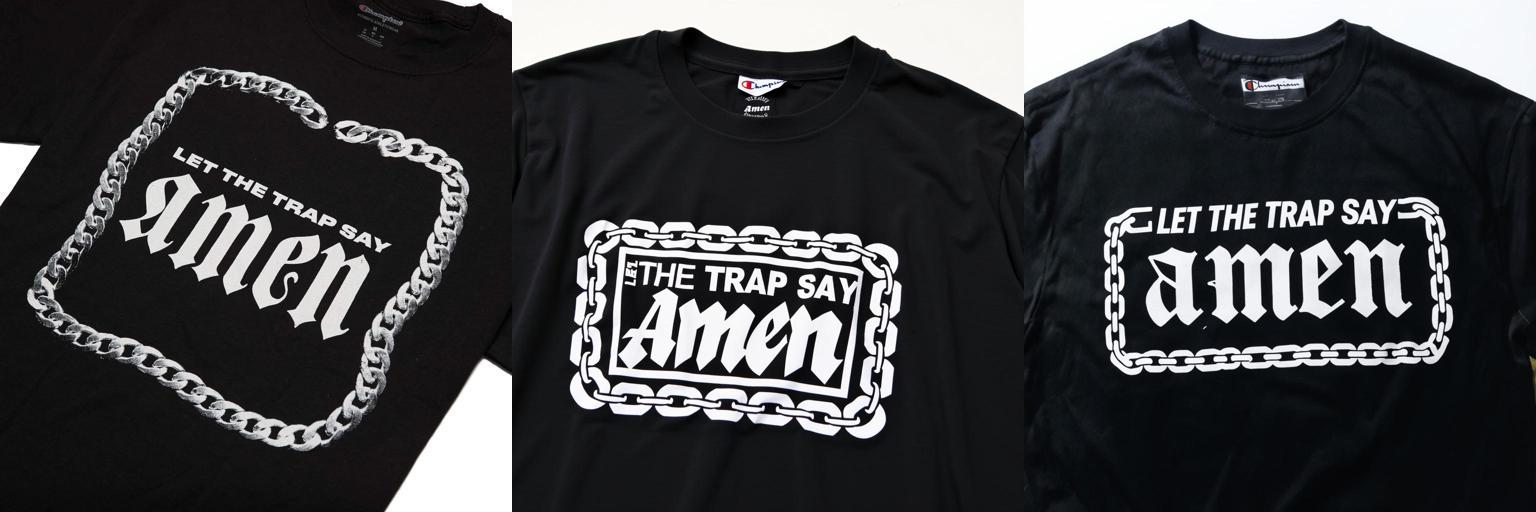}
\begin{tabular}{l}
    \multirow{6}{120mm}{\scriptsize{A close-up photograph of a black T-shirt with a bold and urban design. The T-shirt features a prominent white print in the center. The text '\textbf{LET THE TRAP SAY}' is written in bold, uppercase, sans-serif font and is positioned at the top of the print, slightly tilted. Below this, the word '\textbf{amen}' is displayed in a large, bold, blackletter font. The entire text is outlined by a thick, white, cuban link chain design, which forms a rectangular frame around the text. The T-shirt is made of a plain black fabric, and the neckline label shows '\textbf{Champion}' in small, black, uppercase letters. The background is plain white, providing a stark contrast to the black T-shirt and white print. The lighting is even and soft, highlighting the details of the print without casting harsh shadows.}}
\end{tabular}
\vspace{18mm}

\caption{Qualitative comparison of text rendering accuracy in Mario-text-1k. From left to right: Ground Truth, Flux-pro and Playground V3. PGv3 successfully render the word "LET" and also make sure "amen" is in lower case following the prompt.}

\label{fig:mario}
\end{figure*}

Following previous work~\cite{chen2023textdiffuserdiffusionmodelstext}, we selected 1k samples from Mario-eval, which includes text images in Laion, TMDB and OpenLibrary. We calculated a clip score and used Kosmos 2.5~\cite{lv2024kosmos25multimodalliteratemodel} OCR model to detect text, and calculated text rendering accuracy. In Tab.~\ref{tbl:mario}, our model has much better text accuracy with an OCR-Fscore of 40.35, higher than Flux-pro's 35.28. We also show qualitative examples in Fig.~\ref{fig:mario}, where our model excels at small text.

\begin{table*}[h]
\vspace{-2mm}
\centering
\setlength{\tabcolsep}{6pt}
\renewcommand{\arraystretch}{1.2}
\caption{Mario-text-1k text rendering evaluation benchmark. We generate Flux-dev/pro samples using their official APIs; for SD3-Medium, we used the API from Diffusers. }
\begin{tabular}{lcccc}
\toprule 
    Method & CLIP Score $\uparrow$ & OCR-Precision $\uparrow$& OCR-Recall $\uparrow$& OCR-Fscore $\uparrow$\\ 
    \midrule
    
    SD3-Medium~\cite{esser2024scalingrectifiedflowtransformers} & 31.26 & 15.99 & 21.36 & 18.29 \\
    Flux-dev~\cite{Flux} & 33.41 & 32.86 & 37.38 & 34.98 \\
    Flux-pro~\cite{Flux} & 33.65 & 33.12 & 37.74 & 35.28 \\
    Playground v3 & \textbf{33.89} & \textbf{39.23} & \textbf{41.54} & \textbf{40.35}   \\

\bottomrule
\end{tabular}
\label{tbl:mario}
\end{table*}

\subsection{Evaluation on Image Quality}
We present results on two standard benchmarks, ImageNet and MSCOCO, to compare our model’s performance against other state-of-the-art models. 

However, we maintain our position that ImageNet and MSCOCO may not effectively capture the comprehensive performance of large-scale text-to-image models, especially given the progress of models like PGv3 in prompt-following.

\subsubsection{ImageNet Benchmark}
\begin{table*}[h]
\vspace{-2mm}
\centering
\setlength{\tabcolsep}{6pt}
\renewcommand{\arraystretch}{1.2}
\caption{ImageNet image quality evaluation. All the samples are evaluated at 256 resolution.}
\begin{tabular}{lcccc}
\toprule
    & \multicolumn{2}{c}{train} & \multicolumn{2}{c}{val} \\ 
    Method & FID & FD$_{DINOv2}$ & FID & FD$_{DINOv2}$ \\ 
    \midrule
    \multicolumn{5}{l}{\textit{Class condition models}} \\
    DiT-XL/2~\cite{peebles2023scalablediffusionmodelstransformers} & 9.62 & & & 79.36\\ 
    Simple Diffusion~\cite{hoogeboom2023simple}  & 2.77 & & 3.75 & \\  
    \midrule
    \multicolumn{5}{l}{\textit{Text condition models}} \\
    Playground v2.5~\cite{li2024playground} &  & & 21.18 & 231.05\\ 
    EDM2-xxl-CLIP~\cite{Karras2024edm2} & & & 16.37 & 118.74 \\

    Playground v3 (ours) & & & 14.67 & 102.91 \\
\bottomrule
\end{tabular}
\label{tbl:imagenet}
\end{table*}

In Tab.~\ref{tbl:imagenet}, we compare PGv3 with our previous model PGv2.5. Since our model takes text as a condition, we follow previous work \cite{esser2024scalingrectifiedflowtransformers} to use \textit{a photo of a \{class\}} format to convert ImageNet class labels into text conditions. We also compared a SoTA method EDM2 \cite{Karras2024edm2} model in the same setting. We augmented it with a clip text conditioner for a fair comparison. From the table, PGv3 performs better in terms of FID and FD DINOv2. We also show SoTA class conditioned model performance as a reference; usually, those models take the class label as a condition to achieve better performance. 

\subsubsection{MSCOCO Benchmark}
\begin{table*}[h]
\vspace{-2mm}
\centering
\setlength{\tabcolsep}{6pt}
\renewcommand{\arraystretch}{1.2}
\caption{MSCOCO evaluation. Here we compare our previous method PG2.5 at resolutions 256 and 512, and SoTA open-source methods in 1024 resolution. We compare with original caption (usually short) and PG caption (longer with more details). }
\begin{tabular}{lcccc}
\toprule
    & \multicolumn{2}{c}{Original caption} & \multicolumn{2}{c}{PG Captioner caption} \\ 
    Method & FID & FD$_{DINOv2}$ & FID & FD$_{DINOv2}$ \\ 
    \midrule
    \multicolumn{5}{l}{\textbf{256 $\times$ 256 resolution}} \\
    Playground v2.5~\cite{li2024playground} & 11.41 & 225.80& 8.64& 191.17\\ 
    Playground v3 (ours) & 11.94 & \textbf{172.34}& \textbf{7.06} & \textbf{58.82}\\  
    \midrule
    \multicolumn{5}{l}{\textbf{512 $\times$ 512 resolution}} \\
    Playground v2.5~\cite{li2024playground} & 10.63& 182.00& 14.16& 180.78\\ 
    Playground v3 (ours) & 13.79& \textbf{160.84}& \textbf{7.14}& \textbf{58.41}\\
    \midrule
    \multicolumn{5}{l}{\textbf{1024 $\times$ 1024 resolution}} \\
    PixArt-Sigma~\cite{chen2024pixartsigmaweaktostrongtrainingdiffusion} & 21.60 & 291.29& 15.29& 192.58 \\
    Hunyuan-DiT~\cite{li2024hunyuan} & 22.97 & 339.42& 15.98& 237.73\\
    AuraFlow-0.2~\cite{AuraFlow} & 22.29 & 314.66 & 15.79 & 189.91 \\
    Lumina-Next~\cite{zhuo2024luminanextmakingluminat2xstronger} & 19.04& 342.93& 15.32& 243.16\\
    SD3-Medium~\cite{esser2024scalingrectifiedflowtransformers} & 19.77& 194.97& 14.42& 128.28\\

    Flux-dev~\cite{Flux} & 22.25 & 281.77 & 13.54 & 124.83 \\
    Playground v3 (ours) & \textbf{18.38}& 238.07 & \textbf{8.58}& \textbf{82.59}\\
\bottomrule
\end{tabular}
\label{tbl:mscoco}
\end{table*}

We also evaluated our models in MSCOCO validation set with 30k prompts. In table~\ref{tbl:mscoco} we evaluated both the original prompt (short) and also a reception prompt using PG Captioner (long). We measure the model performance in FID and FD DINOv2~\cite{oquab2023dinov2}, we found FD DINOv2 is a better metrics that is more sensitive to object shapes. We measured our model at different training phases, and compared with the PGv2.5 model for 256 and 512 resolutions. We also compared other SoTA models at the 1024$\times$1024 resolution.

From the table, our model achieves great performance in both short and long prompts and show significant improvement in FD DINOv2 which indicates much better geometry and object shape. Here, we only compare to SoTA open-source models as this evaluation needs to sample 30k images for both the original caption and captions from PG Captioner. From the results, we can see PGv3 achieves the best FID and FD DINOv2 for both caption sets. Notably with PG Captioner captions, our 1024px model significantly outperforms Flux-dev, indicating strong image quality in terms of color and shape generalization. 

\subsection{Evaluation on VAE Reconstruction}
\begin{table*}[h]
\vspace{-2mm}
\centering
\setlength{\tabcolsep}{6pt}
\renewcommand{\arraystretch}{1.2}
\caption{VAE reconstruction evaluation.}
\begin{tabular}{lccc}
\toprule
    Method & LPIPS $\downarrow$ & SSIM $\uparrow$& PSNR $\uparrow$\\ 
    \midrule
    \multicolumn{4}{l}{\textbf{256 $\times$ 256 resolution}} \\
    SDXL-vae-ch4 & 0.116 & 0.797& 24.96 \\ 
    PGv3-vae-ch16 & 0.075 & 0.914& 29.76\\  
    \midrule
    \multicolumn{4}{l}{\textbf{512 $\times$ 512 resolution}} \\
    SDXL-vae-ch4 & 0.112 & 0.821& 26.52 \\ 
    PGv3-vae-ch16 & 0.068 & 0.926& 31.60 \\ 
    \midrule
    \multicolumn{4}{l}{\textbf{1024 $\times$ 1024 resolution}} \\
    SDXL-vae-ch4 & 0.116 & 0.840& 28.11 \\ 
    PGv3-vae-ch16 & 0.060 & 0.939& 33.44 \\ 
\bottomrule
\end{tabular}
\label{tbl:vae-rec}
\end{table*}

Following previous works \cite{dai2023emu, esser2024scalingrectifiedflowtransformers}, we increased the VAE latent channel size from 4 to 16, and we trained the VAE in a progressive manner from lower resolution to higher resolution. In Tab.~\ref{tbl:vae-rec}, we report reconstruction performance on a validation set on multiple resolutions. From the table, we improve significantly compared to SDXL channel 4 VAE; specifically, at resolution 1024x1024, we achieve 33.44 PSNR. 

\section{Captioning Evaluation}
\label{sec:caption_eval}

We used CapsBench to evaluate PG Captioner, along with leading proprietary models GPT-4o and Claude-3.5 Sonnet. When generating captions with proprietary models, we used detailed instructions with output schema, few-shot prompting (by providing three examples of high-quality detailed captions) and chain-of-thought reasoning (perform written analysis of the image prior to generating the caption) to achieve the best results.

\begin{figure*}[t]
\centering
\vspace{-5mm}
\includegraphics[width=0.7\textwidth]{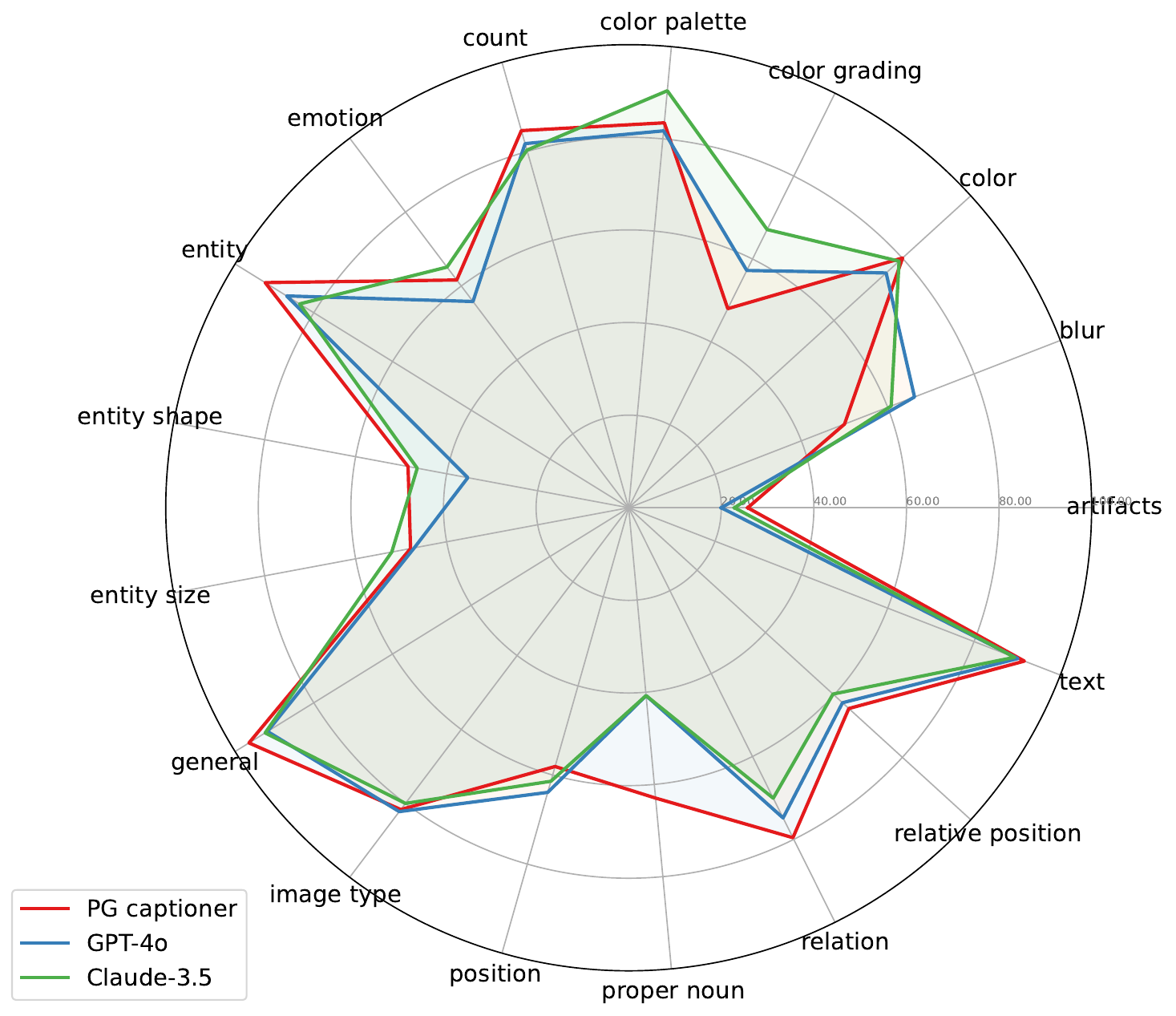}
\caption{Captioning evaluation results split by category}
\label{fig:captioner_bench_results}
\end{figure*}

For question answering, we used Claude-3.5 Sonnet. To work around stochastic outputs of Sonnet and ensure consistency, we run question answering three times for every caption and use the consensus answer. As a result, combined score of PG Captioner is 72.19\%, Claude-3.5 Sonnet's 71.78\%, GPT-4o's 70.66\%. Fig.~\ref{fig:captioner_bench_results} shows the results split by categories. We can see that CapsBench helps to reveal shortcomings of existing models for captioning. Captions usually don't contain descriptions of object shapes (e.g., they likely produce "brown wooden table," instead of "brown rectangular wooden table"), sizes in the image (e.g., "the man appears big in the image"), and visual artifacts (like noise or lens flare). This may be a problem for using these captions in systems heavily reliant on completeness of the captions, such as image generation models or systems for visually impaired users.

\begin{figure*}[t]
\centering
\vspace{-5mm}
\includegraphics[width=0.7\textwidth]{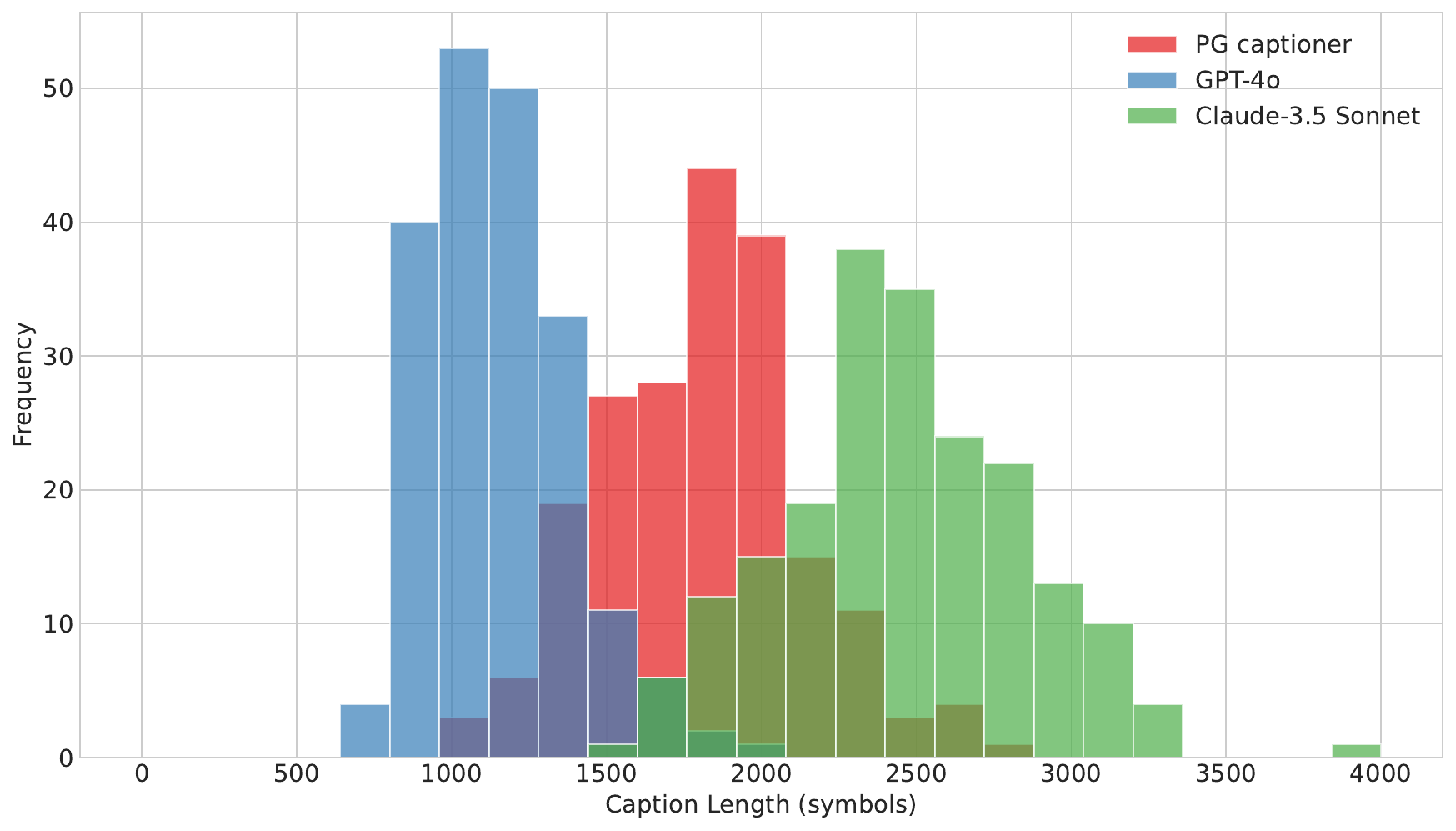}
\caption{Caption length distributions}
\label{fig:caption_lengths}
\end{figure*}

Additionally, we evaluated the length of the captions to understand whether longer captions correlate with higher evaluation scores. Fig.~\ref{fig:caption_lengths} shows histograms of caption lengths for evaluated models. Interestingly, all three models have different length distributions, where GPT-4o generates the shortest captions and Claude-3.5 Sonnet produces captions that are twice as long on average.

\begin{figure*}[t]
\centering
\vspace{-5mm}
\includegraphics[width=1.0\textwidth]{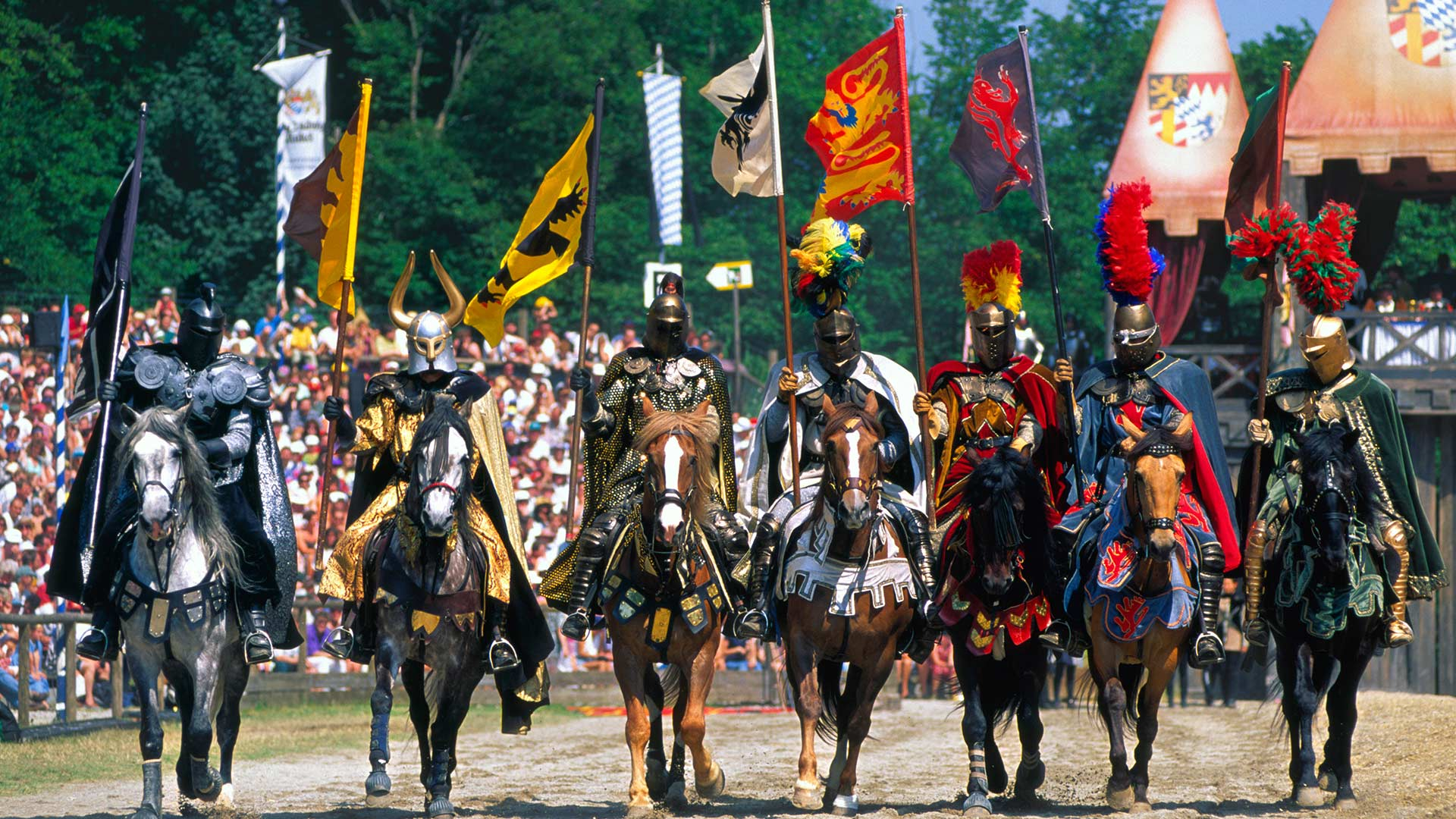}
\caption{Example image from CapsBench}
\label{fig:caption_image_example}
\end{figure*}

To further study the pros and cons of different models, we performed qualitative analysis of captions. In Fig.~\ref{fig:caption_image_example}, we show one of the images from CapsBench. This image contains a significant number of subjects and a lot of small details, which makes it very challenging for VLMs. Tab.~\ref{tbl:capsbench_questions} shows the questions for this image along with the answers generated from the captions. Captions themselves are presented below. 

We can see that PG Captioner describes each knight and their horses in detail. Interestingly, the correctness of descriptions decreases as the model goes from the left knight to the right. PG Captioner also hallucinates the text on the flag in the background, which is not visible due to blur effects. Sonnet, on the other hand, doesn't provide detailed description of each knight and describes them as a group. It also provides more detailed description of the crowd and trees in the background, which makes the caption noticeably longer. GPT-4o generates very concise caption compared to other models. It also describes the knights as a group and incorrectly describes the building on the right as multiple tents.

\begin{table*}[h]
\vspace{-2mm}
\centering
\setlength{\tabcolsep}{6pt}
\renewcommand{\arraystretch}{1.2}
\caption{Questions for the example image from CapsBench}
\begin{tabular}{lp{6cm}cccc}
\toprule 
    Category & Question & GT & PG & Claude & GPT-4o \\ 
    \midrule
    color & Is there a golden yellow flag? & yes & yes & n/a & n/a \\
    color & Is the cape of the right-most knight emerald green? & yes & no & n/a & n/a \\
    position & Are all the knights positioned in the center of the image? & yes & n/a & yes & yes \\
    relative position & Is the knight with the golden horns to the left of the knight with the red feathers? & yes & yes & n/a & n/a \\
    blur & Is the background blurred? & yes & no & no & no \\
    entity size & Is the building on the right side of the image large? & yes & n/a & n/a & n/a \\
    count & Are there seven knights in the image? & yes & yes & yes & yes \\
    entity & Is there a horse in the image? & yes & yes & yes & yes \\
    relation & Is the knight with red feathers mounted on a horse? & yes & yes & n/a & n/a \\
    general & Does the image show an outdoor scene? & yes & yes & yes & yes \\
    image type & Is the image a photograph? & yes & yes & yes & yes \\
    color grading & Does the image have high saturation? & yes & n/a & yes & n/a \\
\bottomrule
\end{tabular}
\label{tbl:capsbench_questions}
\end{table*}

\textbf{PG Captioner caption (accuracy 58.3\%)}: \textit{"A vibrant photograph captures seven knights riding horses through an outdoor medieval reenactment event, positioned in front of a large crowd and tall trees under clear blue skies. The scene is bathed in natural daylight, creating high contrast between the brightly colored costumes and the verdant surroundings. Each knight wears ornate armor and carries distinctive flags with various crests and emblems, including symbols such as lions, dragons, eagles, and heraldic designs. From left to right, the first knight rides a white horse adorned with black and gold armor and a cape featuring circular patterns, holding a black flag with a red lion emblem on a wooden staff. Next is a knight on a gray horse wearing golden armor with intricate designs and a horned silver helmet, carrying a yellow flag with a brown eagle. The third knight, mounted on a brown horse with a white blaze, dons black chainmail armor with gold accents and a white cape with small triangular cut-outs, bearing a white banner displaying a black eagle or dragon silhouette. The fourth knight rides a dark brown steed with a white blaze, clad in elaborate red and gold armor with a flowing crimson cloak, holding a red flag with golden dragon motifs. The fifth knight, atop a light brown horse with a white blaze, sports navy blue armor with red and green trimmings and a forest green cape, bearing a purple flag with a red lion design. The sixth knight, mounted on a black horse, wears metallic silver armor with prominent shoulder plates, a closed visor helmet, and a cape with intricate embroidery, carrying a black flag with a white stag or bull symbol. The seventh knight, on a black horse, completes the lineup with charcoal gray armor embellished with brass details, a unique horn-like helmet, and a cape featuring alternating vertical stripes of royal blue and forest green, topped with a large feather plume in vivid red, yellow, and green hues. Behind them, spectators fill wooden benches and stand areas, attentively watching the procession. Large banners with various designs and text are visible among the crowd, one reading 'Deutscher Ritterturnier' in white serif font on a blue and white striped background. In the background, tall trees provide a lush backdrop, while large tents with conical roofs display colorful shields with various heraldry symbols, including blue and white triangles over red lines on beige backgrounds. One tent features a shield with red and white stripes. The overall mood is festive, celebratory, and historical, capturing the essence of a lively medieval renaissance festival where the past meets present in a picturesque setting."}

\textbf{Claude-3.5 Sonnet caption (accuracy 50\%)}: \textit{"This photograph captures a grand medieval-themed event, likely a renaissance fair or historical reenactment. The image is dominated by a symmetrical arrangement of seven knights on horseback, positioned in the foreground of a sunny outdoor arena. Each knight is fully outfitted in elaborate medieval armor, complete with helmets, plumes, and colorful regalia that spans a vibrant spectrum from silver and gold to rich reds, blues, and greens. The knights are mounted on well-groomed horses of various colors - white, brown, and black - each adorned with decorative barding that matches their riders' colors and heraldry. The knights proudly display an array of flags and banners, each bearing unique heraldic designs such as dragons, lions, and geometric patterns. These banners add vertical elements to the composition and infuse the scene with a sense of pageantry and historical authenticity. Behind the knights, a vast crowd of spectators fills the background, suggesting this is a popular and well-attended event. The audience creates a sea of indistinct faces and colors, providing depth to the image. Beyond the crowd, a line of green trees is visible, framing the upper portion of the photograph and adding a natural backdrop to the medieval spectacle. On the right side of the image, a partial view of what appears to be a medieval-style structure can be seen. It has a conical roof and might be a decorative tower or large tent, further enhancing the period atmosphere of the event. The lighting is bright and natural, indicative of a sunny day, which allows for high detail visibility and vivid color reproduction. This lighting emphasizes the metallic shine of the armor, the rich hues of the fabrics, and the overall festive mood of the occasion. The composition is centered and symmetrical, drawing the viewer's eye to the line of knights as the main focal point. The image has a high level of saturation, bringing out the bold colors of the costumes and flags, while maintaining good contrast that helps distinguish the intricate details of the armor and heraldic designs. Overall, this photograph captures a moment of historical reenactment that combines educational value with entertainment, showcasing the pageantry and spectacle of medieval tournaments and celebrations. The scale of the event, evident from the large crowd and elaborate costumes, suggests a significant and popular cultural gathering celebrating historical heritage and craftsmanship."}

\textbf{GPT-4o caption (accuracy 41.7\%)}: \textit{"The image captures a medieval-themed performance or reenactment, featuring seven knights in full, elaborate medieval armor riding horses. The photograph is taken from a front angle, directly facing the knights, with a well-exposed, bright natural lighting highlighting the scene. The knights are central to the composition, positioned in a symmetrical row. Each knight is dressed in highly detailed armor, including full-body armaments with intricate designs, and each carries a different colored flag adorned with unique heraldic symbols. The knights are mounted on variously colored horses, with some in white, brown, and black. Their helmets are unique, showcasing different designs and colorful plumes, adding to the dramatic and historical mood of the image. The midground displays a large audience dressed in modern-day clothing, attentively watching the performance. They are situated behind a secured viewing area, bordered by fencing. Several colorful banners and flags are visible in the midground, attached to poles and adding to the festive and vibrant atmosphere. The background features lush green trees, giving a natural backdrop to the scene, with medieval-styled tents visible in the far corners, completing the historical setting. The lighting emphasizes the bright, vivid colors and the intricate details of both the knights' armor and their surroundings. The overall composition, lighting, and vivid colors create a lively and historically immersive scene, capturing the essence of a medieval reenactment event."}
\vspace{-3mm}
\section{Conclusion}
\vspace{-2mm}
In this work, we introduce Playground v3, a state-of-the-art text-to-image generation model with strong prompt adherence, reasoning and text rendering ability. It shows superhuman graphic design abilities from a user preference study on various downstream design applications, where we compare samples generated by our model and a high quality curated image dataset created by human designers. We also release to the community a new carefully prepared image captioning benchmark \textit{CapsBench}, designed for detailed long captions, to facilitate the research and benchmarking of the progress of image understanding ability.

\clearpage
{\small
\bibliographystyle{plain}
\bibliography{egbib}

\begin{thebibliography}{10}

\bibitem{anderson2016spice}
Peter Anderson, Basura Fernando, Mark Johnson, and Stephen Gould.
\newblock Spice: Semantic propositional image caption evaluation.
\newblock In {\em Computer Vision--ECCV 2016: 14th European Conference, Amsterdam, The Netherlands, October 11-14, 2016, Proceedings, Part V 14}, pages 382--398. Springer, 2016.

\bibitem{banerjee2005meteor}
Satanjeev Banerjee and Alon Lavie.
\newblock Meteor: An automatic metric for mt evaluation with improved correlation with human judgments.
\newblock In {\em Proceedings of the acl workshop on intrinsic and extrinsic evaluation measures for machine translation and/or summarization}, pages 65--72, 2005.

\bibitem{behnamghader2024llm2vec}
Parishad BehnamGhader, Vaibhav Adlakha, Marius Mosbach, Dzmitry Bahdanau, Nicolas Chapados, and Siva Reddy.
\newblock Llm2vec: Large language models are secretly powerful text encoders.
\newblock {\em arXiv preprint arXiv:2404.05961}, 2024.

\bibitem{betker2023improving}
James Betker, Gabriel Goh, Li~Jing, Tim Brooks, Jianfeng Wang, Linjie Li, Long Ouyang, Juntang Zhuang, Joyce Lee, Yufei Guo, et~al.
\newblock Improving image generation with better captions.
\newblock {\em Computer Science. https://cdn. openai. com/papers/dall-e-3. pdf}, 2(3):8, 2023.

\bibitem{beyer2024paligemma}
Lucas Beyer, Andreas Steiner, Andr{\'e}~Susano Pinto, Alexander Kolesnikov, Xiao Wang, Daniel Salz, Maxim Neumann, Ibrahim Alabdulmohsin, Michael Tschannen, Emanuele Bugliarello, et~al.
\newblock Paligemma: A versatile 3b vlm for transfer.
\newblock {\em arXiv preprint arXiv:2407.07726}, 2024.

\bibitem{chen2023textdiffuserdiffusionmodelstext}
Jingye Chen, Yupan Huang, Tengchao Lv, Lei Cui, Qifeng Chen, and Furu Wei.
\newblock Textdiffuser: Diffusion models as text painters, 2023.

\bibitem{chen2024pixartsigmaweaktostrongtrainingdiffusion}
Junsong Chen, Chongjian Ge, Enze Xie, Yue Wu, Lewei Yao, Xiaozhe Ren, Zhongdao Wang, Ping Luo, Huchuan Lu, and Zhenguo Li.
\newblock Pixart-sigma: Weak-to-strong training of diffusion transformer for 4k text-to-image generation, 2024.

\bibitem{chen2023pixart}
Junsong Chen, Jincheng Yu, Chongjian Ge, Lewei Yao, Enze Xie, Yue Wu, Zhongdao Wang, James Kwok, Ping Luo, Huchuan Lu, et~al.
\newblock Pixart-alpha: Fast training of diffusion transformer for photorealistic text-to-image synthesis.
\newblock {\em arXiv preprint arXiv:2310.00426}, 2023.

\bibitem{chen2023pali}
Xi~Chen, Xiao Wang, Lucas Beyer, Alexander Kolesnikov, Jialin Wu, Paul Voigtlaender, Basil Mustafa, Sebastian Goodman, Ibrahim Alabdulmohsin, Piotr Padlewski, et~al.
\newblock Pali-3 vision language models: Smaller, faster, stronger.
\newblock {\em arXiv preprint arXiv:2310.09199}, 2023.

\bibitem{Cho2024DSG}
Jaemin Cho, Yushi Hu, Jason Baldridge, Roopal Garg, Peter Anderson, Ranjay Krishna, Mohit Bansal, Jordi Pont-Tuset, and Su~Wang.
\newblock Davidsonian scene graph: Improving reliability in fine-grained evaluation for text-to-image generation.
\newblock In {\em ICLR}, 2024.

\bibitem{dai2023emu}
Xiaoliang Dai, Ji~Hou, Chih-Yao Ma, Sam Tsai, Jialiang Wang, Rui Wang, Peizhao Zhang, Simon Vandenhende, Xiaofang Wang, Abhimanyu Dubey, Matthew Yu, Abhishek Kadian, Filip Radenovic, Dhruv Mahajan, Kunpeng Li, Yue Zhao, Vladan Petrovic, Mitesh~Kumar Singh, Simran Motwani, Yi~Wen, Yiwen Song, Roshan Sumbaly, Vignesh Ramanathan, Zijian He, Peter Vajda, and Devi Parikh.
\newblock Emu: Enhancing image generation models using photogenic needles in a haystack, 2023.

\bibitem{dar2022analyzing}
Guy Dar, Mor Geva, Ankit Gupta, and Jonathan Berant.
\newblock Analyzing transformers in embedding space.
\newblock {\em arXiv preprint arXiv:2209.02535}, 2022.

\bibitem{dubey2024llama}
Abhimanyu Dubey, Abhinav Jauhri, Abhinav Pandey, Abhishek Kadian, Ahmad Al-Dahle, Aiesha Letman, Akhil Mathur, Alan Schelten, Amy Yang, Angela Fan, et~al.
\newblock The llama 3 herd of models.
\newblock {\em arXiv preprint arXiv:2407.21783}, 2024.

\bibitem{durrani2020analyzing}
Nadir Durrani, Hassan Sajjad, Fahim Dalvi, and Yonatan Belinkov.
\newblock Analyzing individual neurons in pre-trained language models.
\newblock {\em arXiv preprint arXiv:2010.02695}, 2020.

\bibitem{esser2024scalingrectifiedflowtransformers}
Patrick Esser, Sumith Kulal, Andreas Blattmann, Rahim Entezari, Jonas Müller, Harry Saini, Yam Levi, Dominik Lorenz, Axel Sauer, Frederic Boesel, Dustin Podell, Tim Dockhorn, Zion English, Kyle Lacey, Alex Goodwin, Yannik Marek, and Robin Rombach.
\newblock Scaling rectified flow transformers for high-resolution image synthesis, 2024.

\bibitem{AuraFlow}
FAL.
\newblock Auraflow.

\bibitem{fang2024vila}
Yunhao Fang, Ligeng Zhu, Yao Lu, Yan Wang, Pavlo Molchanov, Jang~Hyun Cho, Marco Pavone, Song Han, and Hongxu Yin.
\newblock Vila2: Vila augmented vila.
\newblock {\em arXiv preprint arXiv:2407.17453}, 2024.

\bibitem{ghosh2023genevalobjectfocusedframeworkevaluating}
Dhruba Ghosh, Hanna Hajishirzi, and Ludwig Schmidt.
\newblock Geneval: An object-focused framework for evaluating text-to-image alignment, 2023.

\bibitem{he2024mars}
Wanggui He, Siming Fu, Mushui Liu, Xierui Wang, Wenyi Xiao, Fangxun Shu, Yi~Wang, Lei Zhang, Zhelun Yu, Haoyuan Li, et~al.
\newblock Mars: Mixture of auto-regressive models for fine-grained text-to-image synthesis.
\newblock {\em arXiv preprint arXiv:2407.07614}, 2024.

\bibitem{hessel2022clipscore}
Jack Hessel, Ari Holtzman, Maxwell Forbes, Ronan~Le Bras, and Yejin Choi.
\newblock Clipscore: A reference-free evaluation metric for image captioning, 2022.

\bibitem{hong2024cogvlm2}
Wenyi Hong, Weihan Wang, Ming Ding, Wenmeng Yu, Qingsong Lv, Yan Wang, Yean Cheng, Shiyu Huang, Junhui Ji, Zhao Xue, et~al.
\newblock Cogvlm2: Visual language models for image and video understanding.
\newblock {\em arXiv preprint arXiv:2408.16500}, 2024.

\bibitem{hoogeboom2023simple}
Emiel Hoogeboom, Jonathan Heek, and Tim Salimans.
\newblock Simple diffusion: End-to-end diffusion for high resolution images, 2023.

\bibitem{hu2023infometic}
Anwen Hu, Shizhe Chen, Liang Zhang, and Qin Jin.
\newblock Infometic: An informative metric for reference-free image caption evaluation.
\newblock {\em arXiv preprint arXiv:2305.06002}, 2023.

\bibitem{hu2024ella}
Xiwei Hu, Rui Wang, Yixiao Fang, Bin Fu, Pei Cheng, and Gang Yu.
\newblock Ella: Equip diffusion models with llm for enhanced semantic alignment.
\newblock {\em arXiv preprint arXiv:2403.05135}, 2024.

\bibitem{Ideogram}
Ideogram.
\newblock Ideogram.

\bibitem{jiang2019tiger}
Ming Jiang, Qiuyuan Huang, Lei Zhang, Xin Wang, Pengchuan Zhang, Zhe Gan, Jana Diesner, and Jianfeng Gao.
\newblock Tiger: Text-to-image grounding for image caption evaluation.
\newblock {\em arXiv preprint arXiv:1909.02050}, 2019.

\bibitem{karras2022elucidating}
Tero Karras, Miika Aittala, Timo Aila, and Samuli Laine.
\newblock Elucidating the design space of diffusion-based generative models, 2022.

\bibitem{Karras2024edm2}
Tero Karras, Miika Aittala, Jaakko Lehtinen, Janne Hellsten, Timo Aila, and Samuli Laine.
\newblock Analyzing and improving the training dynamics of diffusion models.
\newblock In {\em Proc. CVPR}, 2024.

\bibitem{kingma2014adam}
Diederik~P Kingma.
\newblock Adam: A method for stochastic optimization.
\newblock {\em arXiv preprint arXiv:1412.6980}, 2014.

\bibitem{kingma2013auto}
DP~Kingma.
\newblock Auto-encoding variational bayes.
\newblock {\em arXiv preprint arXiv:1312.6114}, 2013.

\bibitem{Flux}
Black~Forest Labs.
\newblock Flux.

\bibitem{lee2021qace}
Hwanhee Lee, Thomas Scialom, Seunghyun Yoon, Franck Dernoncourt, and Kyomin Jung.
\newblock Qace: Asking questions to evaluate an image caption.
\newblock {\em arXiv preprint arXiv:2108.12560}, 2021.

\bibitem{li2024playground}
Daiqing Li, Aleks Kamko, Ehsan Akhgari, Ali Sabet, Linmiao Xu, and Suhail Doshi.
\newblock Playground v2.5: Three insights towards enhancing aesthetic quality in text-to-image generation, 2024.

\bibitem{li2023blip}
Junnan Li, Dongxu Li, Silvio Savarese, and Steven Hoi.
\newblock Blip-2: Bootstrapping language-image pre-training with frozen image encoders and large language models.
\newblock In {\em International conference on machine learning}, pages 19730--19742. PMLR, 2023.

\bibitem{li2024hunyuan}
Zhimin Li, Jianwei Zhang, Qin Lin, Jiangfeng Xiong, Yanxin Long, Xinchi Deng, Yingfang Zhang, Xingchao Liu, Minbin Huang, Zedong Xiao, et~al.
\newblock Hunyuan-dit: A powerful multi-resolution diffusion transformer with fine-grained chinese understanding.
\newblock {\em arXiv preprint arXiv:2405.08748}, 2024.

\bibitem{lipman2022flow}
Yaron Lipman, Ricky~TQ Chen, Heli Ben-Hamu, Maximilian Nickel, and Matt Le.
\newblock Flow matching for generative modeling.
\newblock {\em arXiv preprint arXiv:2210.02747}, 2022.

\bibitem{liu2024improved}
Haotian Liu, Chunyuan Li, Yuheng Li, and Yong~Jae Lee.
\newblock Improved baselines with visual instruction tuning.
\newblock In {\em Proceedings of the IEEE/CVF Conference on Computer Vision and Pattern Recognition}, pages 26296--26306, 2024.

\bibitem{liu2024llm4gen}
Mushui Liu, Yuhang Ma, Xinfeng Zhang, Yang Zhen, Zeng Zhao, Zhipeng Hu, Bai Liu, and Changjie Fan.
\newblock Llm4gen: Leveraging semantic representation of llms for text-to-image generation.
\newblock {\em arXiv preprint arXiv:2407.00737}, 2024.

\bibitem{liu2022flow}
Xingchao Liu, Chengyue Gong, and Qiang Liu.
\newblock Flow straight and fast: Learning to generate and transfer data with rectified flow.
\newblock {\em arXiv preprint arXiv:2209.03003}, 2022.

\bibitem{lv2024kosmos25multimodalliteratemodel}
Tengchao Lv, Yupan Huang, Jingye Chen, Yuzhong Zhao, Yilin Jia, Lei Cui, Shuming Ma, Yaoyao Chang, Shaohan Huang, Wenhui Wang, Li~Dong, Weiyao Luo, Shaoxiang Wu, Guoxin Wang, Cha Zhang, and Furu Wei.
\newblock Kosmos-2.5: A multimodal literate model, 2024.

\bibitem{nichol2021glide}
Alex Nichol, Prafulla Dhariwal, Aditya Ramesh, Pranav Shyam, Pamela Mishkin, Bob McGrew, Ilya Sutskever, and Mark Chen.
\newblock Glide: Towards photorealistic image generation and editing with text-guided diffusion models.
\newblock {\em arXiv preprint arXiv:2112.10741}, 2021.

\bibitem{oquab2023dinov2}
Maxime Oquab, Timoth{\'e}e Darcet, Th{\'e}o Moutakanni, Huy Vo, Marc Szafraniec, Vasil Khalidov, Pierre Fernandez, Daniel Haziza, Francisco Massa, Alaaeldin El-Nouby, et~al.
\newblock Dinov2: Learning robust visual features without supervision.
\newblock {\em arXiv preprint arXiv:2304.07193}, 2023.

\bibitem{papineni2002bleu}
Kishore Papineni, Salim Roukos, Todd Ward, and Wei-Jing Zhu.
\newblock Bleu: a method for automatic evaluation of machine translation.
\newblock In {\em Proceedings of the 40th annual meeting of the Association for Computational Linguistics}, pages 311--318, 2002.

\bibitem{peebles2023scalable}
William Peebles and Saining Xie.
\newblock Scalable diffusion models with transformers, 2023.

\bibitem{peebles2023scalablediffusionmodelstransformers}
William Peebles and Saining Xie.
\newblock Scalable diffusion models with transformers, 2023.

\bibitem{podell2023sdxl}
Dustin Podell, Zion English, Kyle Lacey, Andreas Blattmann, Tim Dockhorn, Jonas Müller, Joe Penna, and Robin Rombach.
\newblock Sdxl: Improving latent diffusion models for high-resolution image synthesis, 2023.

\bibitem{qin2024diffusiongpt}
Jie Qin, Jie Wu, Weifeng Chen, Yuxi Ren, Huixia Li, Hefeng Wu, Xuefeng Xiao, Rui Wang, and Shilei Wen.
\newblock Diffusiongpt: Llm-driven text-to-image generation system.
\newblock {\em arXiv preprint arXiv:2401.10061}, 2024.

\bibitem{radford2021learning}
Alec Radford, Jong~Wook Kim, Chris Hallacy, Aditya Ramesh, Gabriel Goh, Sandhini Agarwal, Girish Sastry, Amanda Askell, Pamela Mishkin, Jack Clark, et~al.
\newblock Learning transferable visual models from natural language supervision.
\newblock In {\em International conference on machine learning}, pages 8748--8763. PMLR, 2021.

\bibitem{raffel2020exploring}
Colin Raffel, Noam Shazeer, Adam Roberts, Katherine Lee, Sharan Narang, Michael Matena, Yanqi Zhou, Wei Li, and Peter~J Liu.
\newblock Exploring the limits of transfer learning with a unified text-to-text transformer.
\newblock {\em Journal of machine learning research}, 21(140):1--67, 2020.

\bibitem{ramesh2022hierarchical}
Aditya Ramesh, Prafulla Dhariwal, Alex Nichol, Casey Chu, and Mark Chen.
\newblock Hierarchical text-conditional image generation with clip latents.
\newblock {\em arXiv preprint arXiv:2204.06125}, 1(2):3, 2022.

\bibitem{Rombach_2022_CVPR}
Robin Rombach, Andreas Blattmann, Dominik Lorenz, Patrick Esser, and Bj\"orn Ommer.
\newblock High-resolution image synthesis with latent diffusion models.
\newblock In {\em Proceedings of the IEEE/CVF Conference on Computer Vision and Pattern Recognition (CVPR)}, pages 10684--10695, June 2022.

\bibitem{rombach2022highresolution}
Robin Rombach, Andreas Blattmann, Dominik Lorenz, Patrick Esser, and Björn Ommer.
\newblock High-resolution image synthesis with latent diffusion models, 2022.

\bibitem{ronneberger2015u}
Olaf Ronneberger, Philipp Fischer, and Thomas Brox.
\newblock U-net: Convolutional networks for biomedical image segmentation.
\newblock In {\em Medical image computing and computer-assisted intervention--MICCAI 2015: 18th international conference, Munich, Germany, October 5-9, 2015, proceedings, part III 18}, pages 234--241. Springer, 2015.

\bibitem{saharia2022photorealistictexttoimagediffusionmodels}
Chitwan Saharia, William Chan, Saurabh Saxena, Lala Li, Jay Whang, Emily Denton, Seyed Kamyar~Seyed Ghasemipour, Burcu~Karagol Ayan, S.~Sara Mahdavi, Rapha~Gontijo Lopes, Tim Salimans, Jonathan Ho, David~J Fleet, and Mohammad Norouzi.
\newblock Photorealistic text-to-image diffusion models with deep language understanding, 2022.

\bibitem{sharifzadeh2024synth}
Sahand Sharifzadeh, Christos Kaplanis, Shreya Pathak, Dharshan Kumaran, Anastasija Ilic, Jovana Mitrovic, Charles Blundell, and Andrea Banino.
\newblock Synth2: Boosting visual-language models with synthetic captions and image embeddings.
\newblock {\em arXiv preprint arXiv:2403.07750}, 2024.

\bibitem{song2024moma}
Kunpeng Song, Yizhe Zhu, Bingchen Liu, Qing Yan, Ahmed Elgammal, and Xiao Yang.
\newblock Moma: Multimodal llm adapter for fast personalized image generation.
\newblock {\em arXiv preprint arXiv:2404.05674}, 2024.

\bibitem{su2024roformer}
Jianlin Su, Murtadha Ahmed, Yu~Lu, Shengfeng Pan, Wen Bo, and Yunfeng Liu.
\newblock Roformer: Enhanced transformer with rotary position embedding.
\newblock {\em Neurocomputing}, 568:127063, 2024.

\bibitem{tian2024u}
Yuchuan Tian, Zhijun Tu, Hanting Chen, Jie Hu, Chao Xu, and Yunhe Wang.
\newblock U-dits: Downsample tokens in u-shaped diffusion transformers.
\newblock {\em arXiv preprint arXiv:2405.02730}, 2024.

\bibitem{vedantam2015cider}
Ramakrishna Vedantam, C~Lawrence~Zitnick, and Devi Parikh.
\newblock Cider: Consensus-based image description evaluation.
\newblock In {\em Proceedings of the IEEE conference on computer vision and pattern recognition}, pages 4566--4575, 2015.

\bibitem{wang2021faier}
Sijin Wang, Ziwei Yao, Ruiping Wang, Zhongqin Wu, and Xilin Chen.
\newblock Faier: Fidelity and adequacy ensured image caption evaluation.
\newblock In {\em Proceedings of the IEEE/CVF Conference on Computer Vision and Pattern Recognition}, pages 14050--14059, 2021.

\bibitem{xiao2024florence}
Bin Xiao, Haiping Wu, Weijian Xu, Xiyang Dai, Houdong Hu, Yumao Lu, Michael Zeng, Ce~Liu, and Lu~Yuan.
\newblock Florence-2: Advancing a unified representation for a variety of vision tasks.
\newblock In {\em Proceedings of the IEEE/CVF Conference on Computer Vision and Pattern Recognition}, pages 4818--4829, 2024.

\bibitem{zhuo2024luminanextmakingluminat2xstronger}
Le~Zhuo, Ruoyi Du, Han Xiao, Yangguang Li, Dongyang Liu, Rongjie Huang, Wenze Liu, Lirui Zhao, Fu-Yun Wang, Zhanyu Ma, Xu~Luo, Zehan Wang, Kaipeng Zhang, Xiangyang Zhu, Si~Liu, Xiangyu Yue, Dingning Liu, Wanli Ouyang, Ziwei Liu, Yu~Qiao, Hongsheng Li, and Peng Gao.
\newblock Lumina-next: Making lumina-t2x stronger and faster with next-dit, 2024.

\end{thebibliography}
}

\end{document}